\documentclass{article}

\usepackage{arxiv}

\usepackage[utf8]{inputenc} 
\usepackage[T1]{fontenc}    
\usepackage{hyperref}       
\usepackage{url}            
\usepackage{booktabs}       
\usepackage{amsfonts}       
\usepackage{nicefrac}       
\usepackage{microtype}      
\usepackage{lipsum}		
\usepackage{graphicx}
\usepackage{natbib}
\usepackage{doi}
\usepackage{tikz-cd}
\usepackage{mathtools}

\usepackage{amsmath}
\usepackage{amssymb}
\usepackage{amsthm}
\usepackage[boxruled]{algorithm2e}

\newtheorem{theorem}{Theorem}
\newtheorem{definition}{Definition}
\newtheorem{lemma}{Lemma}

\title{Universal Decision Models}

\author{ Sridhar Mahadevan \thanks{Draft under revision. Comments welcome}\\
	Adobe Research and University of Massachusetts, Amherst\\
	\texttt{smahadev@adobe.com, mahadeva@umass.edu} \\
	
}

\renewcommand{\shorttitle}{\textit{arXiv} Template}

\hypersetup{
pdftitle={A template for the arxiv style},
pdfsubject={q-bio.NC, q-bio.QM},
pdfauthor={David S.~Hippocampus, Elias D.~Striatum},
pdfkeywords={First keyword, Second keyword, More},
}

\begin{document}
\maketitle

\begin{abstract}
Humans are universal decision makers: we reason causally to understand the world; we act competitively to gain advantage in commerce, games, and war; and we are able to learn to make better decisions through trial and error. Whilst these individual modalities of decision making have been studied for decades in various subfields of AI and ML, there has been commensurately less effort in developing formalisms that unify these various modalities into common framework. In this paper, we propose Universal Decision Model (UDM), a mathematical formalism based on category theory, to address this challenge. Decision objects in a UDM correspond to instances of decision tasks, ranging from  causal models and dynamical systems  such as Markov decision processes and predictive state representations,  to network multiplayer games and  Witsenhausen's intrinsic models, which generalizes all these previous formalisms. A UDM is a category of objects, which include decision objects, observation objects, and solution objects.  Bisimulation morphisms map between decision objects that capture  structure-preserving abstractions. We formulate universal properties of UDMs, including information integration,  decision solvability,  and hierarchical abstraction.  Information integration consolidates data from heterogeneous sources by forming products or limits in the UDM category.  Abstraction simulates complex decision processes by simpler processes through bisimulation morphisms by forming quotients, co-products and co-limits in the UDM category. Finally, solvability of a UDM decision object is defined by a fixed point equation, and it corresponds to an isotonic order-preserving morphism across the topology induced by UDM objects. We describe universal functorial representations of UDMs, and propose an algorithm for computing the minimal object in a UDM using algebraic topology.  We sketch out an application of UDMs to causal inference in network economics, using a complex multiplayer producer-consumer two-sided marketplace.

\end{abstract}


\section{Introduction}

One of the singular aspects of human cognition is our universal decision capacity: we reason causally to interact with and understand the world from a young age \citep{DBLP:journals/cogsci/SobelTG04}, and continue to do so into adulthood \citep{pearl-book,rubin-book}. We act competitively when it benefits us in arms control negotiations, commerce, and games \citep{game-theory-book,mas-book}. Since we almost always make sub-optimal decisions, due to incomplete information and computational limitations \citep{DBLP:journals/jair/RussellS95},  we  learn to make better decisions through trial and error \citep{DBLP:books/lib/SuttonB98}. Whilst these individual decision making modalities have been studied for decades in AI \citep{DBLP:books/aw/RN2020}, we nonetheless possess  an inadequate understanding of how to  integrate these disparate abilities. It appears we understand the parts of universal decision making far better than the whole!  The main contribution of this paper is a novel theory of universal decision making, codified in a mathematical framework we call Universal Decision Model (UDM). The bulk of the paper is focused on understanding the {\em information structures} that guide decision making. In particular, our paper does not specifically address the algorithmic aspects of universal decision making, although we touch upon this topic towards the end. Furthermore, optimization plays a central role in much of the literature in {\em sequential}  decision making. As the scope of UDM is far broader than sequential decision making, which is a very specialized information structure, optimization as traditionally conceived plays only a minor role in the UDM framework. 

Our paper is also motivated by the growing need to understand how to scale existing formalisms to extremely large and complex decision systems, both to understand complex behavior in biology, and to control large decentralized computing systems.  Our work is related to category-theoretic models of complex interconnected systems \citep{fong2016algebra}. Consider a group of computing elements that form a cloud AI implementation, which are tasked to make decisions on gathering and processing data from a heterogeneous set of sources \citep{DBLP:series/synthesis/2020Lin}. Similarly, consider the challenge faced by a group of honeybees that are scouting their environment for a new location for their hive \citep{honeybee-democracy}. Our framework provides a fresh perspective on these challenges, bringing a powerful formalism of categorial thinking to shed light on universal properties underlying decision making in these different realms. Our approach does not assume  any a priori ordering on the agent structure, which must be discovered or designed to make the problem feasible. Each agent may be unaware if it should act first or last, or indeed, if it should act at all. The organization of agents into a linearly or partially ordered structure may vary, depending on the state of nature, randomness in observations, or the task at hand. To ensure unique solvability of complex decision making tasks by such large organizations of agents, the fundamental {\em information structures} that underlie decision making must be carefully designed.

\section{Universal Decision Model: Informal Overview}

\begin{figure}[h]
\begin{center} \hfill
\begin{minipage}{0.95\textwidth}
\includegraphics[scale=1.2]{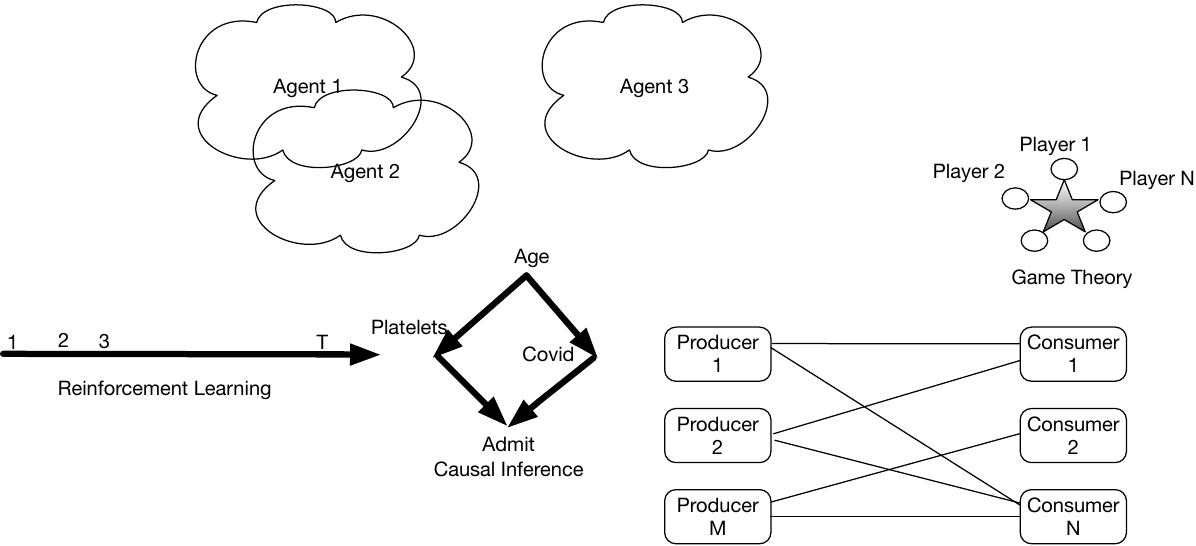}
\end{minipage}
\end{center}
\caption{Universal decision model (UDM) is a unifying framework that integrates decision making across multiple modalities, from reinforcement learning  (left) and causal inference (middle) to  complex network games (right).}
\label{if}
\end{figure}

Figure~\ref{if} illustrates the broad Universal Decision Model (UDM) framework studied in this paper, which seeks to elucidate the common information structures that underly a variety of decision making modalities that have been extensively studied in a broad swath of literature, including AI \citep{DBLP:books/aw/RN2020}, control theory \citep{witsenhausen:75}, game theory \citep{game-theory-book,algorithmic-game-theory} statistics \citep{rubin-book}, psychology \cite{DBLP:journals/cogsci/SobelTG04}, and network economics \cite{nagurney:vibook}. Broadly, a UDM involves a collection of {\em elements} $A$ (representing agents, causal variables, points in time, economic entities etc.), a decision space $U_\alpha \in A$ for each ``actor" $\alpha \in A$ that has an associated measurable space $(U_\alpha, {\cal F}_\alpha)$ (over which a suitable probability space can be defined), and most importantly, an {\em information field} \citep{DBLP:journals/mst/Witsenhausen73} that represents each agent's state of knowledge regarding its decision.  Each agent $\alpha$ makes a decision using a policy $\pi_\alpha: \prod_\beta U_\beta \rightarrow U_\alpha$ that defines a measurable function over $({\cal I}_\alpha, {\cal F}_\alpha)$. The measurability condition imposes an abstract constraint on what an agent knows in making a decision. At the one extreme,  if the agent's information field ${\cal I}_\alpha \subset {\cal F}(\emptyset)$, that implies it can act without depending on any of the other decision makers. More generally, the information field ${\cal I}_\alpha \subset {\cal F}(B)$ for some subset $B \subset A$ of decision makers, which imposes a (pre or partial) ordering on the decision makers. 

\subsection{Two Real-World Applications} 
 
 \begin{figure}[h]
\begin{center} \hfill
\begin{minipage}{0.5\textwidth}
\includegraphics[scale=0.1]{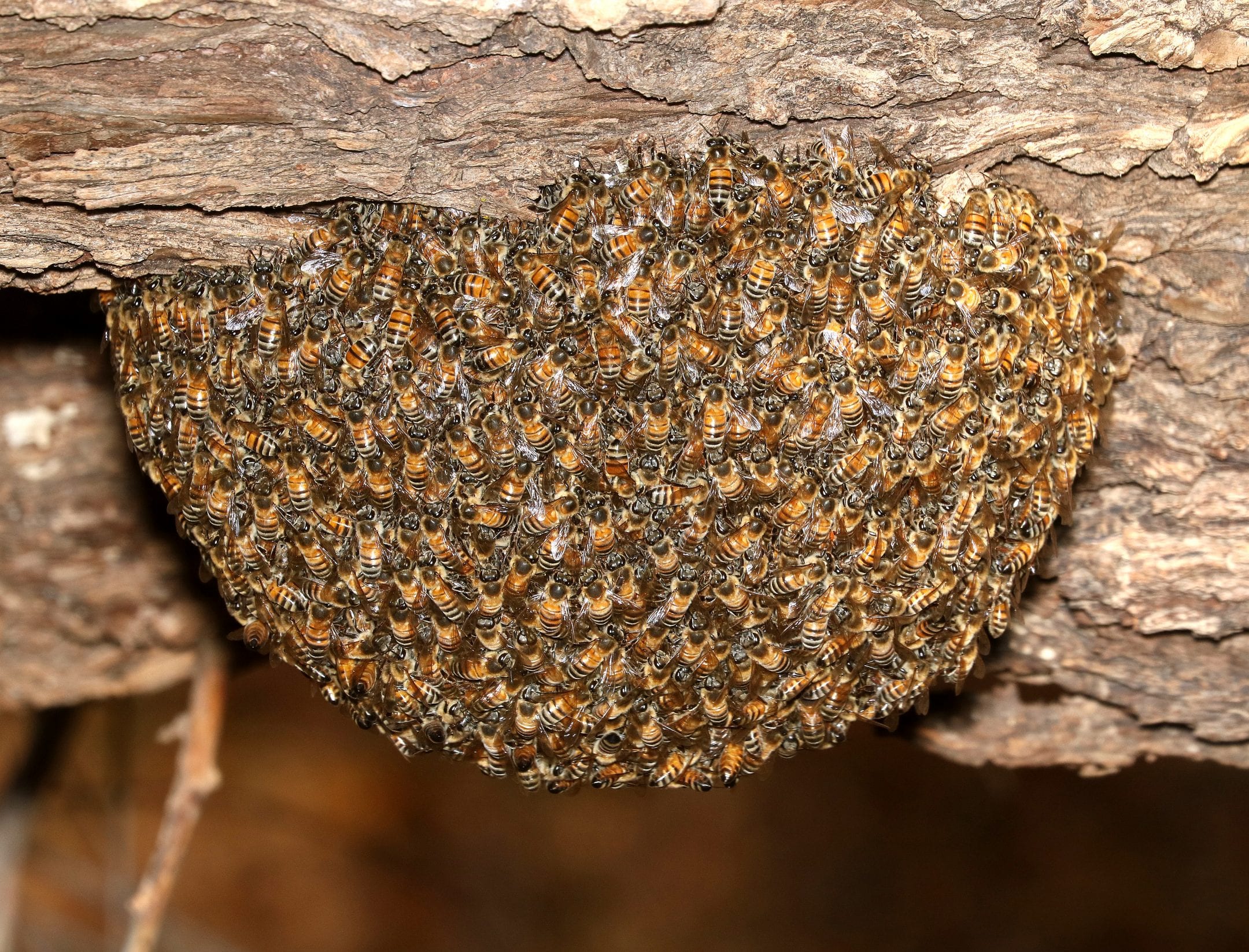}
\end{minipage} \hspace{0.5in}
\begin{minipage}{0.3\textwidth}
\includegraphics[scale=0.2]{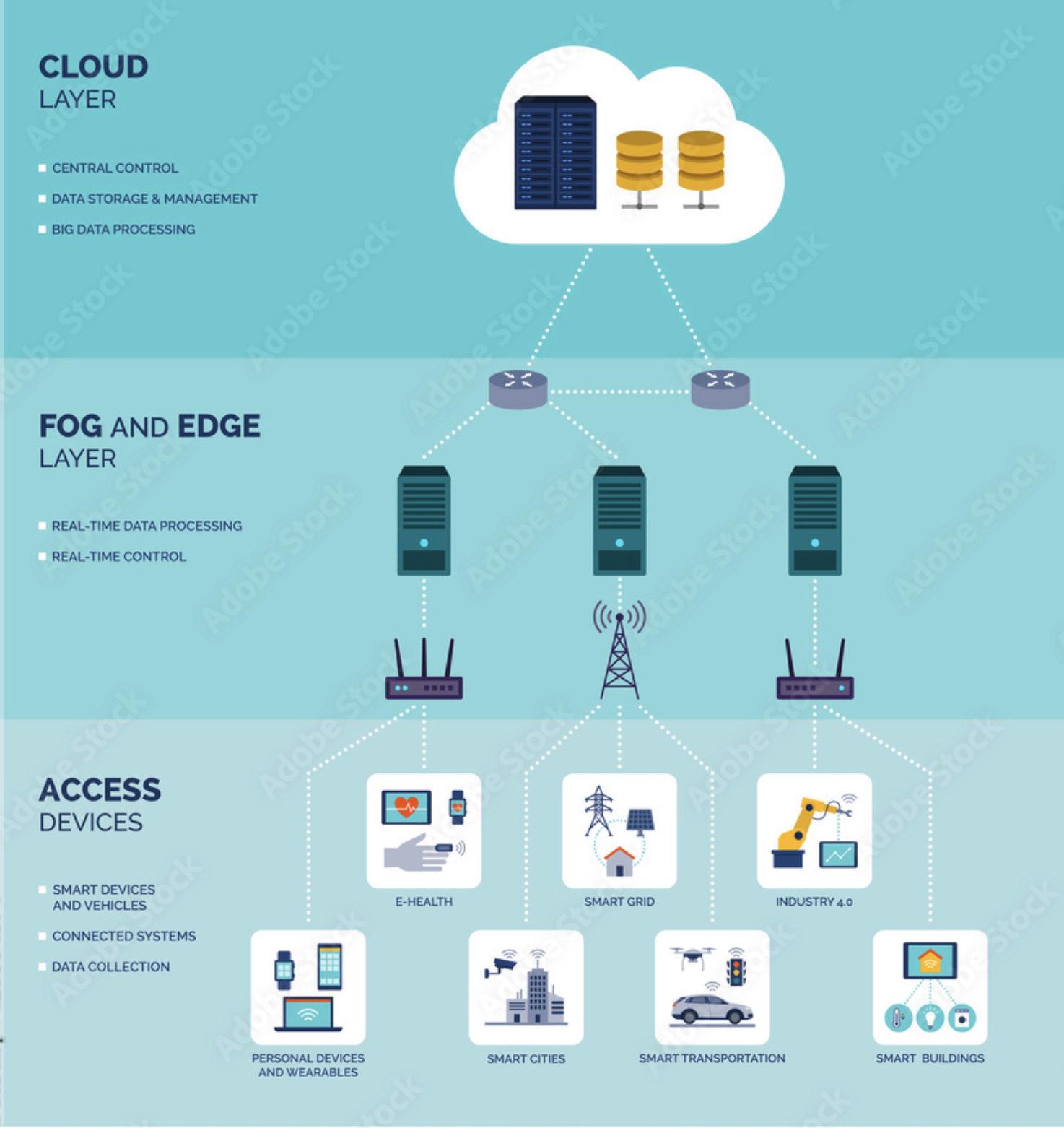}
\end{minipage}
\end{center}
\caption{Left: a swarm of honeybees scouting for a new location are required to make a complex life-altering decision based on collecting information from  surveillance flights \citep{honeybee-democracy}. Right: A generic cloud computing network can be decomposed into subsystems, comprised of computing nodes with varying degrees of information access.}
\label{edge}
\end{figure}

To motivate the following theoretical development, we turn to two practical applications, one involving the design of cloud computing systems, and the second involving the computation of equilibria in network economics. Together, these real-world problems will illustrate the need to develop more sophisticated notions of agency that the usual formalisms in causal inference, game theory and RL currently enable.

Figure~\ref{edge} illustrates challenges of decision making by complex groups of agents in biology and in technology. A swarm of thousands of honeybees are required to make a life-altering decision on where to relocate their hive based on reconnaissance flights, and lack of any a priori fixed coordination mechanism among the bees \citep{honeybee-democracy}. A generic cloud computing network, which is organized into subsystems with varying levels of information storage, compute power, and responsiveness. In our paper, we abstract from the specifics of such systems, and in fact, even potential applications of such networks. Our focus is primarily on understanding how to theoretically characterize the {\em information structures} underlying such systems. For example, computing elements at the lowest level of the cloud network may be individual IOT devices or smartphones. These have visibility into data collected at an individual level. In contrast, the subsystems at the fog or edge layer have greater visibility at the aggregate population level, but due to privacy concerns, may have access to only aggregate statistics of individual data.

\subsection{Information Fields} 

Our work draws extensively on the idea of {\em information fields}, a key component of Witsenhausen's intrinsic model  \citep{causality-Witsenhausen71,DBLP:journals/iandc/Witsenhausen71,DBLP:journals/mst/Witsenhausen73,DBLP:journals/mcss/Witsenhausen88}. Information fields provide a foundation to analyze decision making in a wide range of settings, from game playing, to decentralized decision making and multiagent stochastic control. A book length treatment of the intrinsic model is given in \citep{carpentier:hal-01165572}. The intrinsic model continues to be studied \citep{DBLP:reference/sc/Grover15,DBLP:journals/tac/NayyarMT11,DBLP:journals/tac/NayyarMT13,DBLP:journals/tac/NayyarT19,DBLP:conf/cdc/NayyarB12}, and was recently shown to generalize Pearl's causal do-calculus \citep{DBLP:journals/corr/heymann}. 

 At its core, the {\em intrinsic model} is based on a measure-theoretic approach for representing  {\em information fields} -- the data available to make a decision at some point.  Witsenhausen introduced the notion of a {\em subsystem} that defines a topology on the finite space of entities in the model, based on a (reflexive, transitive) {\em pre-ordering} relationships based on each element's information fields. A key insight of his is the discovery that the subsystem relationship is intimately related to the nature of the overall decision-making problem. For example, a {\em team} decision making problem involves entities that can act without knowledge of each other's information fields, which defines a topology where the subsystems correspond to singletons. In contrast, a $T_0$ (Kolmogorov) topology is defined by a {\em sequential } intrinsic model where there exists a fixed ordering $(\alpha_1, \ldots, \alpha_n)$ of the (decision makers, variables) entities such that the information field for entity $\alpha_k$ is purely a function of the fields defined by entities that preceded it in this fixed ordering. 

Our main contribution in this paper is to study the categorial foundations of the intrinsic model, namely elucidate the universal properties of information fields that underlie complex decision making.  We focus on three universal properties: information integration, decision solvability and hierarchical abstraction. An agent is constantly required to make decisions given partial information about its environment. The ability to act thus requires integration of information from multiple sources into a sufficient statistic for action. A decision problem must be solvable in a well-defined manner, which almost always can be shown to reduce to solving a fixed point equation. Finally, complex decision problems must be decomposable for decision making to scale: hierarchical abstraction that ignores details is an essential component for scalability.

Our paper uses two fundamental guiding principles. The first principle regards the definition of what is considered a {\em universal property}, which is based on category theory \citep{riehl}. In category theory, objects are characterized not in terms of their internal structure, but rather the interactions they make with other objects. A universal property of an object, consequently, is a functorial representation of its {\em interaction} with other objects, which serves to define the object up to isomorphism. We are thus interested in answering the fundamental question: given a decision-making object, whether a structural causal model or a game or an MDP, how can we functorially characterize its universal properties up to isomorphism? 

 Rather than describe an object by enumerating its elements, such as commonly done in set theory, category theory builds on the principle of describing objects by their interaction with other objects in the category. This principle is embodied in the {\em Yoneda lemma}, one of the deepest and most influential results in category theory. This lemma formalizes precisely the notion that an object $c$ in a category ${\cal C}$ can be completely described (to within an isomorphism) by the set of all morphisms from the other objects in ${\cal C}$ from $c$, or from other objects to $c$. Morphisms in a category are closed under composition, and satisfy an associative property. {\em Functors} are structure-preserving mappings from one category ${\cal C}$ to another category ${\cal D}$, which map objects in ${\cal C}$ to corresponding objects in ${\cal D}$, but also map morphisms $f: c \rightarrow d$ in ${\cal C}$ to corresponding morphisms in ${\cal D}$. 

We characterize the universal properties of information fields that play a foundational role in decision making using the tools of category theory. Information integration corresponds to the ability to form {\em products} of elements. The product object in a category is universal with respect to the property of having unique morphisms to its factors, such that every mapping to one of the factors must be uniquely decomposable through the product. This universal property is shown to underlie a range of decision making formalisms, from decision making to causal inference. Another key notion is that of a {\em bisimulation} between objects representing decision making processes \citep{DBLP:conf/category/ArbibM74,DBLP:conf/lics/JoyalNW93}, which underpins the widespread use of homomorphisms in MDPs and predictive state representations (PSRs) \citep{DBLP:conf/ijcai/RavindranB03,DBLP:conf/aaai/DeanG97,DBLP:conf/aaai/SoniS07}. We characterize bisimulation as a universal property through quotient spaces, defined by an equivalence relation $\sim$, where the quotient space $X / \sim$ is uniquely characterized by the ability to map objects such that isomorphic objects have the same image. We will see that the notion of subsystems in intrinsic models is based on the universal property of quotient spaces, wherein agents form equivalence classes of subsystems based on shared information fields. 

Decision solvability in causal inference, games and reinforcement learning all involve finding fixed points of a system of equations. For example, in recursive structural causal models $M = (U,V,F,P)$, where $U$ is a set of exogenous variables, $V$ is a set of endogenous variables where each $x_i \in V$ is a function $f_i \in F$ of some subset of variables $U \cup V - \{ x_i \}$, and $P$ is a probability over the exogenous variables $U$, recursive solvability implies there is a fixed ordering of the variables $U \cup V$ such that each exogenous variable takes on a unique value $X_i = f(\mbox{Pa}_i)$ (where $\mbox{Pa}_i$ refers to the ``parents" of $x_i$) defined for some particular probability $P(u)$ of the exogenous variables. This constraint imposes a fixed point requirement on solvability. Similarly, in game theory, each agent must be able to compute a best response behavior based on knowledge of the other agents' actions. Finally, in reinforcement learning, the Bellman optimality condition imposes a fixed point solvability constraint. All of these constraints can be shown to follow the general causality principle elucidated by Witsenhausen \citep{causality-Witsenhausen71}. We characterize the universal properties of information fields that induce solvable decision problems in all these specialized settings. 
 
 \subsection{Information Fields}
 
 The concept of information fields had its origins in work on game theory \citep{aumann,aumann-borel,game-theory-book}, and we introduce it first in that setting where it can be described in a simpler way. In general, a group of decision makers only have partial knowledge of the true state of nature, referred to below by a (continuous or discrete) set $\Omega$. At any point where an agent $\alpha$ is contemplating a decision, the true state of the world may be indicated by $\omega_0 \in \Omega$, but the agent may only be able to glimpse the true state of $\omega_0$ with some uncertainty, e.g. knowing it belongs to some partition field of $\Omega$. 
 
 \begin{definition} 
 A partial information game ${\cal G} = \langle A, (\Omega, {\cal B}, P), (U_\alpha, {\cal F}_\alpha)_{\alpha \in A} \rangle $, where $A$ is a finite group of players, $\Omega$ defines the states of nature, ${\cal B}$ is the usual Borel topology of sets closed under complementation and countable unions on $\Omega$, so that $(\Omega, {\cal B}, P)$ forms a probability space. Each player can make decisions from a continuous or discrete set $U_\alpha$, and their knowledge of the true state of the world is defined by ${\cal F}_\alpha$, a partition of $\Omega$. 
 \end{definition}
 
 For example, consider the  ensemble of computing elements in a cloud computing network, or in an network economics problem, such as those shown in Figure~\ref{edge}, by the set $A$. Each computing element $a \in A$ can be thought of as an ``actor" that makes decisions over the measurable space $(U_\alpha, {\cal F}_\alpha)$. Intuitively, this means that $U_\alpha$ could be a discrete or continuous set of choices, and ${\cal F}_\alpha$ is a partition of $\Omega$.  Consider a two unit network with units $\alpha$ and $\beta$, where the parameters of the game are defined as follows: 
 
 \begin{itemize} 
 
 \item $\Omega = \{1, 2, \ldots, 9 \}$, ${\cal B} = 2^\Omega$, $P\{i: i \in \Omega \} = \frac{1}{9}$. 
 
 \item ${\cal F}_\alpha = \{ \{1, 2, 3 \}, \{4, 5, 6 \}, \{7, 8, 9 \} \}$. 
 
  \item ${\cal F}_\beta= \{ \{1, 2, 3, 4\}, \{5, 6, 7, 8\}, \{ 9 \} \}$. 
 
 \end{itemize} 
 
 We would like to be able to update the partitions based on events. For example, if the computing elements observed the event $\{3, 4 \}$, what would their posteriors look like?
 
 \subsubsection{Operations on Information Fields: Join and Meet} 
 
 To update the priors based on evidence, we introduce the join and meet operations on partition fields, and more generally, on $\sigma$-algebras. \citet{carpentier:hal-01165572} contains an extensive discussion of partition fields, and its relation to $\sigma$-algebras. We now give a simple example of working with partition fields, which will later be generalized to $\sigma$-algebras. For agent $\alpha$, its partition field ${\cal F}_\alpha$ contains a partition of the states of nature $\Omega$. We want to introduce two operations on states of knowledge that will useful in the remainder of the paper, namely join and meet. The set of partition fields, or their generalization, $\sigma$-algebras, form a partially ordered set, or even a lattice. This structure naturally allows computing the least upper bound and the greatest lower bound of a set of elements. We will use the join and meet operations to indicate these as follows.
  
  \begin{definition}
  The {\bf meet} of two partition fields ${\cal F}_\alpha \wedge {\cal F}_\beta$ is defined as finest partition refined by both ${\cal F}_\alpha$ and ${\cal F}_\beta$. In contrast, the {\bf join} of two partitions ${\cal F}_\alpha \vee {\cal F}_\beta$ is the coarsest common refinement of ${\cal F}_\alpha$ and ${\cal F}_\beta$. More formally, we say a partition ${\cal F}_\alpha$ is {\bf finer} than another partition ${\cal F}_\gamma$, denoted as ${\cal F}_\gamma \leq {\cal F}_\alpha$,  if every element of ${\cal F}_\gamma$ is included as an element of ${\cal F}_\alpha$. 
  \end{definition}
  
  In other words, to compute the join of two partition fields, we compute the intersection of every component of ${\cal F}_\alpha$ with that of ${\cal F}_\beta$. To compute the meet of two partition fields, we find the smallest set of subsets that can be composed as the union of partition elements from ${\cal F}_\alpha$ and ${\cal F}_\beta$. Using the above simple example of a game, we get: 
  
  \begin{itemize} 

  \item ${\cal F}_\alpha \vee {\cal F}_\beta = \{ \{1, 2, 3 \}, \{4\}, \{5, 6 \}, \{7, 8\}, \{9 \}\}$. 
 
 \item ${\cal F}_\alpha \wedge {\cal F}_\beta = \{ \{1, 2, 3, 4, 5, 6, 7, 8, 9 \} \}$. Note here that there does not exist a smaller subset that can be constructed out of the subsets in both partition fields. 
 
 \end{itemize} 
 
 Consider $\alpha$ and $\beta$ observing the event $A = \{3, 4\}$. In this case, the posteriors for each of them is given as:
 
 \begin{itemize}
     \item ${\cal F}^A_\alpha$ = \{\{1, 2, 3\}, \{4, 5, 6\} \}, leading to its assessment of the probability of the state of nature being $F_\alpha(\omega) = \frac{P(A \cap P_\alpha(\omega))}{F_\alpha(\omega)} = \frac{1}{3}$. 
     
       \item ${\cal F}^B_\alpha$ = \{\{1, 2, 3, 4\}\}, leading to its assessment of the probability of the state of nature being $P_\beta(\omega) = \frac{P(A \cap P_\alpha(\omega))}{P_\alpha(\omega)} = \frac{1}{2}$. 
 \end{itemize}
 
 \citet{aumann} studied an interesting class of problems where the players cannot agree to disagree if they share common knowledge about an event. A classic example of this problem is the ``muddy children problem", where a group of children are told they can go home if their foreheads are muddy. Each child cannot see his or her own forehead, but they can see the other foreheads, and no communication is allowed otherwise. For a group of $N$ children, it turns out the teacher has to repeat the statement ``At least one child has a muddy forehead" before all the children get up to leave the class. This is a simple but insightful example of the problem of reasoning with common knowledge. We return to this topic later in the paper, and pose it again in the context of information fields. 
 
 \subsubsection{Sigma algebras and Cylindrical Extensions} 
 
 We will work more generally with $\sigma$-algebras, but the underlying concepts are similar to partition fields (a detailed comparison of their properties is given in \citep{carpentier:hal-01165572}).  $\sigma$-algebras are defined on the states of nature $\Omega$ as a collection of subsets ${\cal F}$ that are closed under complementation and countable union, which implies closure under intersection as well, and with the restriction that $\Omega \in {\cal F}$.
 
 \begin{definition}
 A {\bf measurable space} $(U, {\cal F})$ is defined as a set $U$ along with a $\sigma$-algebra ${\cal F}$ of subsets of $U$, closed under complementation and (countable) union, along with the constraint that the complete set $U \in {\cal F}$. 
 \end{definition}
 
 Much of our discussion in this paper will be in the context of measurable functions on measurable spaces. 
 
 \begin{definition}
 A {\bf measurable function} $f: U \rightarrow V$ is defined to be any function defined over measurable spaces in its domain and range, namely if $(U, {\cal F}_U)$ is the measurable space over its domain, and $(V, {\cal F}_V)$ is the measurable space over the range, then every pre-image of a measurable set in the range is measurable in the domain, that is $f^{-1}(Y) \in {\cal F}_U, Y \in {\cal F}_V$. 
 \end{definition}
 An important special case is when the $\sigma$-algebras are finite, in which we can use the following theorem. 
 
 \begin{theorem}
 For any finite measurable space $(U, {\cal F})$, its $\sigma$-algebra ${\cal F}$ can be generated purely from a partition of $U = \{P_1, \ldots, P_k \}$, by forming the union of all possible subsets in the partition. That is, for any $X \in {\cal F}$, it follows that $X = \cup_{i \in I} P_i$, where $I \subset \{1, \ldots, k \}$. 
 \end{theorem}
 
 An important application of this theory is defining observations over information fields. We can state the general definition as follows: 
 
 \begin{definition}
 The {\bf smallest $\sigma$-algebra} $\sigma(C)$ generated by a family of sets $C = \{C_1, \ldots, C_k \}$ is defined as the intersection of all $\sigma$-algebras containing $C$. 
 \end{definition}
 
 In particular, given a topological space the smallest $\sigma$-algebra generated by the topology is called the Borel $\sigma$-algebra. 
 
 \begin{definition}
 The {\bf Borel} $\sigma$-algebra is defined as the smallest $\sigma$-algebra defined by the topological space $(X, {\cal O})$. 
 \end{definition}
 
 This leads naturally to the definition of a probability space $(\Omega, {\cal B}, P)$. A detailed definition of probability measures is given in any textbook on measure theory \citep{halmos:book}. 
 
 \begin{definition}
 The {\bf probability space} $(\Omega, {\cal B}, P)$ is defined as a measurable space $(\Omega, {\cal B})$ with a measurable function $P: \Omega \rightarrow (0,1)$, defined over it, such that $P(\Omega) = 1$, $P(A \cup B) = P(A) + P(B)$ for all disjoint events $A, B$, where $P$ is a measurable function. 
 \end{definition}
 
 Consider for simplicity the case when a computing element represents an IOT sensor that can only measure two values, so in this case, $U_\alpha = \{0, 1\}$. We can choose ${\cal F}_\alpha$ in several ways, ranging from the power set or {\em discrete topology} ${\cal F})_\alpha = P(U_\alpha) = \{\emptyset, \{0 \}, \{1 \}, \{ 0, 1 \} \}$ to the {\em indiscrete} topology ${\cal F}_\alpha = \{\emptyset,\{ 0, 1 \} \}$.  Each computing unit also has some awareness of the ``state of nature", which could be represented as a set of noisy measurements of local and/or global information. The state of nature is modeled as a probability space $(\Omega, {\cal B}, P)$, where $\Omega$ is the sample space of events, ${\cal B}$ is a measurable space of subsets of the sample space, which is also endowed with a Borel topology, and $P$ is the probability measure such that $P(\Omega) = 1$. 

We can also use some simple properties of $\sigma$-algebras. 

\begin{itemize} 

\item If $D \subset C \subset B \subset A$, then ${\cal F}_B(D) \subset {\cal F}_B(C), {\cal F}_B(C) \cup {\cal F}_B(D) = {\cal F}_B(C \cup D), {\cal F}_B(C) \vee {\cal F}_B(D) = {\cal F}_B(C \cup D)$, which will be useful below in defining a topology over computing elements that share a common information field. 

\item Note $H_\emptyset = \Omega$, and ${\cal F}_B(\emptyset)$ is defined as the {\em cylindrical extension} of the $\sigma$-algebra ${\cal B}$ over states of nature to $H_B$. In general, the cylindrical extension of a $\sigma$-algebra $\prod_{\alpha \in B} {\cal F}_\alpha$ for a subset $B \subset A$ to all of $A$ is defined as $\prod_{\alpha \in B} {\cal F}_\alpha \times \prod_{\alpha \notin B} \{ \emptyset, {\cal F}_\alpha \} \subset {\cal F}_A(A) \equiv {\cal F}$. In other words, the $\sigma$-algebra for elements $\alpha \in B$ remains the same, whereas for elements $\alpha \notin B$, we use the maximally uninformative $\sigma$-algebra of the indiscrete topology $\{ \emptyset, {\cal F}_\alpha \}$. 

\end{itemize}

\section{Universal Decision Model} 
\label{udm} 

We now proceed to give a more formal introduction to the Universal Decision Model (UDM), which draws extensively on the concepts in category theory \citep{riehl}, as well as Witsenhausen's information field representation \citep{DBLP:journals/iandc/Witsenhausen71}, suitably generalized to the setting of category theory. Accordingly, we first give a brief review of category theory, and then proceeed to describe UDM. Subsequent chapters will explore particular instantations of UDM models in more concrete settings, such as causal inference, stochastic control and reinforcement learning, and network economics. 

\subsection{Category Theory} 

\begin{table}[t]
\begin{center}
\begin{tabular}{|c |c | } \hline 
{\bf Set theory } & {\bf Category theory} \\ \hline 
 set & object \\ 
 subset & subobject \\
 truth values $\{0, 1 \} $ & subobject classifier $\Omega$ \\
power set $P(A) = 2^A$ & power object $P(A) = \Omega^A$ \\ \hline
bijection & isomorphims \\ 
injection & monic arrow \\
surjection & epic arrow \\ \hline
singleton set $\{ * \}$ & terminal object ${\bf 1}$ \\ 
empty set $\emptyset$ & initial object ${\bf 0}$ \\
elements of a set $X$ & morphism $f: {\bf 1} \rightarrow X$ \\
- & non-global element $Y \rightarrow X$ \\ 
- & functors, natural  transformations \\ 
- & limits, colimits, adjunctions \\ \hline
\end{tabular}
\end{center}
\caption{Comparison of notions from set theory and category theory.} 
\label{set-vs-categories}
\end{table}

Over the past 70 odd years, a concerted effort by a large group of mathematicians has resulted in the development of a sweeping unification of large areas of mathematics using category theory \citep{riehl}. Table~\ref{set-vs-categories} compares the basic notions in set theory vs. category theory. 
Briefly, a category is a collection of objects, and a collection of morphisms between pairs of objects, which are closed under composition, satisfy associativity, and include an identity morphism for every object. For example, sets form a category under the standard morphism of functions. Groups, modules, topological spaces and vector spaces all form categories in their own right, with suitable morphisms (e.g, for groups, we use group homomorphisms, and for vector spaces, we use linear transformations). We will illustrate the application of category theory to reinforcement learning by showing that it is relatively straightforward to define categories over MDPs and PSR models, based on the previously defined homomorphisms over these models. We then summarize previous work on open maps over machines, which generalizes these ideas. 

 A broad class of models used in optimal control, reinforcement learning, operations research and system identification can be characterized in terms of categories and the morphisms between them, including Markov decision processes (MDPs) and semi-MDPs \citep{DBLP:books/wi/Puterman94}, predictive state representations (PSRs) \citep{DBLP:conf/aaai/SoniS07} and subspace identification models in system identification \citep{DBLP:journals/automatica/OverscheeM93}, as well as Witsenhausen's intrinisc model of decentralized stochastic control based on information fields \citep{DBLP:journals/mst/Witsenhausen73}. Our work can also be viewed as a generalization of previous abstraction methods, such as {\em homomorphisms} used in model minimization in MDPs and SMDPs \citep{DBLP:conf/aaai/DeanG97,DBLP:conf/ijcai/RavindranB03} and PSR's \citep{DBLP:conf/aaai/SoniS07}, as well as related abstraction models used in algebraic automata theory \citep{DBLP:journals/iandc/HartmanisS62}. 
 
Our presentation will follow the excellent treatments given in \citep{bradley,goldblatt,riehl}.  Intuitively, a category is simply a {\em collection} of {\em objects} $X, Y, \ldots$,  and a collection of {\em morphisms} $f, g, \ldots $, where $f: X \rightarrow Y$ is the morphism whose domain is $X$ and co-domain is $Y$. A basic principle of category theory is that objects have no discernable internal structure, and their identity up to isomorphism is revealed by their interaction with other objects in the category. To take a simple, but illustrative example, consider a set $X$ with $n$ elements. Rather than list the elements of the set, we define it simply as a collection of mappings from the category $1$ to $X$, where $1$ is the category with exactly one object, and one morphism (identity). Each mapping from $1$ to $X$ must by definition pick out one of its elements, and consequently the entire ensemble of elements in $X$ is revealed by the ensemble of mappings from $1$ to $X$. The Yoneda lemma described later generalizes this principle to mappings from an arbitrary category to the category of sets.  Mappings between categories are known as {\em functors}, and will be defined below.

For each pair of morphisms $f, g$, such that the co-domain of $f$ is the same as the domain of $g$, there is a composite morphism $g f$, simply defined as the composition of $g$ and $f$ (where $f$ is applied first, followed by $g$), defined as $g f: X \rightarrow Z$. There are two additional requirements: each object $X$ has associated with it an {\em identity} morphism $1_X: X \rightarrow X$, whose composition with any other morphism $f: X \rightarrow Y$ is defined as $1_Y f = f = f 1_X = f$. The second requirement is {\em associativity}, whereby given morphisms $f: X \rightarrow Y, g: Y \rightarrow Z, h: Z \rightarrow W$, the composite morphism $h g f: X \rightarrow W$ is associative. 

Some examples of categories are illustrated below, which we will refer to in the remainder of the paper. 

\begin{itemize} 

\item {\bf Set}: The canonical example of a category is {\bf Set}, which has as its objects, sets, and morphisms are functions from one set to another. The {\bf Set} category will play a central role in our framework, as it is fundamental to the universal representation constructed by Yoneda embeddings. 

\item {\bf Top:} The category {\bf Top} has topological spaces as its objects, and continuous functions as its morphisms. Recall that a topological space $(X, \Xi)$ consists of a set $X$, and a collection of subsets $\Xi$ of $X$ closed under finite intersection and arbitrary unions. 

\item {\bf Group:} The category {\bf Group} has groups as its objects, and group homomorphisms as its morphisms. 

\item {\bf Graph:} The category {\bf Graph} has graphs (undirected) as its objects, and graph morphisms (mapping vertices to vertices, preserving adjacency properties) as its morphisms. The category {\bf DirGraph} has directed graphs as its objects, and the morphisms must now preserve adjacency as defined by a directed edge. 

\item {\bf Poset:} The category {\bf Poset} has partially ordered sets as its objects and order-preserving functions as its morphisms. 

\item {\bf Meas:} The category {\bf Meas} has measurable spaces as its objects and measurable functions as its morphisms. Recall that a measurable space $(\Omega, {\cal B})$ is defined by a set $\Omega$ and an associated $\sigma$-field of subsets {\cal B} that is closed under complementation, and arbitrary unions and intersections, where the empty set $\emptyset \in {\cal B}$. 

\end{itemize} 

\subsection{Universal Properties} 

A core goal in category theory is to elucidate the universal properties of objects and morphisms. The motivation is understand the essence of what makes a particular concept unique. For example, in set theory, the cartesian product of two sets $A \times B = \{(a,b) | a \in A, b \in B\}$ is simply defined by listing the elements of the set representing the cartesian product. In category theory, a different approach is taken, one that involves articulating the universal property of objects that represent cartesian products, as we will see below.   One of our primary contributions is the categorial formulation of Witsenhausen's intrinsic model. The key principle underlying category theory is {\em universality}: this seemingly simple concept is somewhat difficult to grasp at first glance since some of its definitions involve a deeper definition of terms that we provide in later sections. Intuitively, let us for now consider the universal property of an object to be something that characterizes all morphisms into or out of the object. This philosophy of describing objects in terms of the interactions they make with other objects is a key characteristic of category theory. 

\subsubsection{Quotients:} 

Quotient spaces induced by an equivalence relation $\sim$  play a fundamental role in the UDM framework. Given a category ${\cal C}_{\mbox{UDM}}$ of decision objects objects, the quotient ${\cal C}_{\mbox{UDM}} / \sim$ is the set of equivalence classes in ${\cal C}_{\mbox{UDM}}$, whereby an object $c \in {\cal C}_{\mbox{UDM}}$ is mapped to its equivalence class $[c]_f$ under the function $f$ such that $f(c) = f(d)$ implies $[c] = [d]$. Reflexivity, symmetry, and transitivity easily follow from the definition. The canonical projection $\pi: C \rightarrow C / \sim$ is the unique map sending $c$ to its equivalence class $[c]_f$.  Quotients will play a key role in the UDM framework as we will see below. 

\begin{center}
 \begin{tikzcd}[column sep=small]
& X \arrow[dl] \arrow[dr] & \\
  X / \sim \arrow{rr} &                         & Y
\end{tikzcd}
\end{center} 

The universal property of quotients is indicated in the above diagram whereby any map from object $X$ to object $Y$ that equates equivalent objects is uniquely factorizable through its quotient map, so that $f = g \pi$, and the diagram commutes. Quotients have played a central role in MDP homomorphisms  \citep{DBLP:conf/aaai/DeanG97,DBLP:conf/ijcai/RavindranB03} and PSR's \citep{DBLP:conf/aaai/SoniS07}, as well as related abstraction models used in algebraic automata theory \citep{DBLP:journals/iandc/HartmanisS62}.

\subsubsection{Product:}  

A central motif in much of the literature in decision making is the need to integrate information from multiple sources. In the RL literature, dynamical system  models like MDPs, POMDPs and PSRs typically assume the notion of a {\em state}, which summarizes all the information from the past (or future) that is important for making optimal decisions. In structural causal models, an endogenous variable in the model is a function of exogenous and other endogenous variables, which requires integrating information from all these ``parent" variables. In games, an agent needs to consider the potential responses of all other other actions. All of these involve the fundamental operation of a product. In category theory, products are defined as the following universal property: 

\begin{center}
\begin{tikzcd}
  T
  \arrow[drr, bend left, "x"]
  \arrow[ddr, bend right, "y"]
  \arrow[dr, dotted, "r" description] & & \\
    & X  \times Y \arrow[r, "p"] \arrow[d, "q"]
      & X \arrow[d, "f"] \\
& Y \arrow[r, "g"] &Z
\end{tikzcd}
\end{center} 

The above figure shows a diagram, a standard construct in category theory, where objects are depicted by vertices with labels, and morphisms are indicated by labeled edges. This diagram asserts that there is an object labeled $X \times Y$ with morphisms $p: X \times Y \rightarrow X$ and $q: X \times Y \rightarrow Y$, which we recognize immediately as the canonical projection from a cartesian product to its components. Furthermore, the diagram asserts that given any morphism from an object $T$ to $X$, there is in fact a unique way to factor that morphism through the product object, so that the diagram ``commutes", meaning the morphism $x = p \ r$. Similarly, any morphism from $T$ to $Y$ is also uniquely factored through $r$, so that $y = q \ r$. We have thus characterized the product object purely in terms of the morphisms into and out of the object.  Our first claim is that for decision making, the ability to form products is a universal property, which is an essential ingredient in any framework.  In the UDM framework, products play a key role in defining information fields, which are a subfield of the product space $(\prod_\alpha U_\alpha, \prod_\alpha {\cal F}_\alpha)$. As we will see later, the ability to form products is essential in using information fields to specify structural causal models, as well as define states in sequential models. 

\subsubsection{Co-Product:}

A related universal property to product is the {\em coproduct} property, which loosely translates to forming ``disjoint" unions of sets. Coproducts refer to the universal property of abstracting a group of elements into a larger one. For example, information fields of multiple decision objects can be combined into one larger information field through co-products. 

\begin{center}
\begin{tikzcd}
    & Z\arrow[r, "p"] \arrow[d, "q"]
      & X \arrow[d, "f"] \arrow[ddr, bend left, "h"]\\
& Y \arrow[r, "g"] \arrow[drr, bend right, "i"] &X \sqcup Y \arrow[dr, "r"]  \\ 
& & & R 
\end{tikzcd}
\end{center} 

In the commutative diagram above, the coproduct object $X \sqcup Y$ uniquely factorizes any mapping $h: X \rightarrow R$ and any mapping $i: Y \rightarrow R$, so that $h = r \ f$, and furthermore $i = r \ g$.

\subsubsection{Pullback and Pushforward Mappings} 

\begin{figure}[h]
\centering
\begin{tikzcd}
  T
  \arrow[drr, bend left, "x"]
  \arrow[ddr, bend right, "y"]
  \arrow[dr, dotted, "k" description] & & \\
    & 
    U\arrow[r, "g'"] \arrow[d, "f'"]
      & X \arrow[d, "f"] \\
& Y \arrow[r, "g"] &Z
\end{tikzcd}
\caption{Universal Property of pullbacks and pushforward mappings.} 
\label{univpr}
\end{figure}
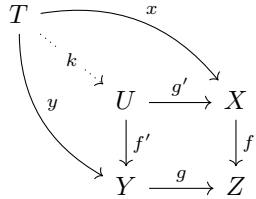

Figure~\ref{univpr}  illustrates the fundamental property of a {\em pullback}, which along with {\em pushforward}, is one of the core ideas in category theory. The pullback square with the objects $U,X, Y$ and $Z$ implies that the composite mappings $g \ f'$ must equal $g' \ f$. In this example, the morphisms $f$ and $g$ represent a {\em pullback} pair, as they share a common co-domain $Z$. The pair of morphisms $f', g'$ emanating from $U$ define a {\em cone}, because the pullback square ``commutes" appropriately. Thus, the pullback of the pair of morphisms $f, g$ with the common co-domain $Z$ is the pair of morphisms $f', g'$ with common domain $U$. Furthermore, to satisfy the universal property, given another pair of morphisms $x, y$ with common domain $T$, there must exist another morphism $k: T \rightarrow U$ that ``factorizes" $x, y$ appropriately, so that the composite morphisms $f' \ k = y$ and $g' \ k = x$. Here, $T$ and $U$ are referred to as {\em cones}, where $U$ is the limit of the set of all cones ``above" $Z$. If we reverse arrow directions appropriately, we get the corresponding notion of pushforward. So, in this example, the pair of morphisms $f', g'$ that share a common domain represent a pushforward pair.

\subsection{Universal Decision Model}

 Now, we introduce the Universal Decision Model (UDM) more formally. In the UDM category ${\cal C}_{\mbox{UDM}}$, as in any category, we are given a collection of {\em decision objects} ${\cal D}$, and a set of morphisms ${\cal M}_{\mbox{UDM}}$ between UDM objects, where $f: c \rightarrow d$ is a morphism that maps from UDM object $c$ to $d$. A morphism need not exist between every pair of UDM objects. In this paper, we restrict ourselves to {\em locally small} UDM categories, meaning that exists only a set's worth of morphisms between any pair of UDM objects. More general categories of UDMs are beyond the scope of this introductory paper. 
 
 \begin{definition} 
 \label{udm-defn}
 A Universal Decision Model (UDM)  is defined as a category ${\cal C}_{\mbox{UDM}}$, where each decision object is represented as a  tuple $\langle (A, (\Omega, {\cal B}, P), U_\alpha, {\cal F}_\alpha, {\cal I}_\alpha)_{\alpha \in A} \rangle$, where $A$ describes a finite universe of elements (e.g., random variable in a structural causal model, dynamical systems, such as linear dynamical systems, MDPs, PSRs etc., intrinsic models, or multiplayer network games), $(\Omega, {\cal B}, P)$ is a probability space representing the inherent stochastic state of nature due to randomness,  $U_\alpha$ is a measurable space from which a decision $u \in U_\alpha$ is chosen by decision object $\alpha$. Each element's policy in a  decision object is any function $\pi_\alpha: \prod_\beta U_\beta \rightarrow U_\alpha$ that is measurable from its information field ${\cal I}_\alpha$, a subfield of  the overall product space $(\prod_\alpha U_\alpha, \prod_\alpha {\cal F}_\alpha)$, to the $\sigma$-algebra ${\cal F}_\alpha$. The policy of decision object $\alpha$ can be any function $\pi_\alpha: \prod_\beta U_\beta \rightarrow U_\alpha$. 
\end{definition}

A UDM may also contain {\em observation objects} and {\em solution objects}, which we discuss later in the paper. Briefly, observation objects correspond to a ``run-time" trace behavior of a decision object, whereas a solution object represents a ``solution" of the decision problem. As mentioned at the outset, the traditional role of optimization in much of (sequential) decision making plays only a minor role in the UDM framework, as it is tailored to a particular information structure. We will discuss solution methodologies for particular information structures later in the paper. 

\begin{definition} 
The {\bf  information field} of an element $\alpha \in A$ in a decision object $c$ in UDM category ${\cal C}_{\mbox{UDM}}$ is denoted as ${\cal I}_\alpha \subset {\cal F}_A(A)$ characterizes the information available to decision object $\alpha$ for choosing a decision $u \in U_\alpha$. 
\end{definition} 

As we will see below, the information field structure yields a surprisingly rich topological space that has many important consequences for how to organize the  decision makers in a complex organization into subsystems. An element $\alpha$ in a decision object requires information from other elements or subsystems in the network. To formalize this notion, we use product decision fields and product $\sigma$-algebras, with their canonical projections.

\begin{definition} 
  Given a subset of nodes $B \subset A$, let $H_B = \Omega \times \prod_{\alpha \in B} U_\alpha$ be the {\bf product space of decisions} of nodes in the subset $B$, where the {\bf product $\sigma$-algebra} is ${\cal B} \times \prod_{\alpha \in B} {\cal F}_\alpha = {\cal F}_B(B)$. It is common to also denote the product $\sigma$-algebra by the notation $\otimes_{\alpha \in A} {\cal F}_\alpha$.  If $C \subset B$, then the {\bf induced $\sigma$-algebra} ${\cal F}_B(C)$ is a subfield of ${\cal F}_B(B)$, which can also be viewed as the inverse image of ${\cal F}_C(C)$ under the canonical projection of $H_B$ onto $H_C$. \footnote{Note that for any cartesian product of sets $\prod_i X_i$, we are always able to uniquely define  a projection map into any component set $X_i$, which is a special case of the product universal property in a category.} 
 \end{definition}
 
 \subsection{Bisimulation Morphims as Open Maps} 
 
\begin{figure}[t]
\begin{center}
\begin{minipage}{0.5\textwidth}
\includegraphics[scale=0.4]{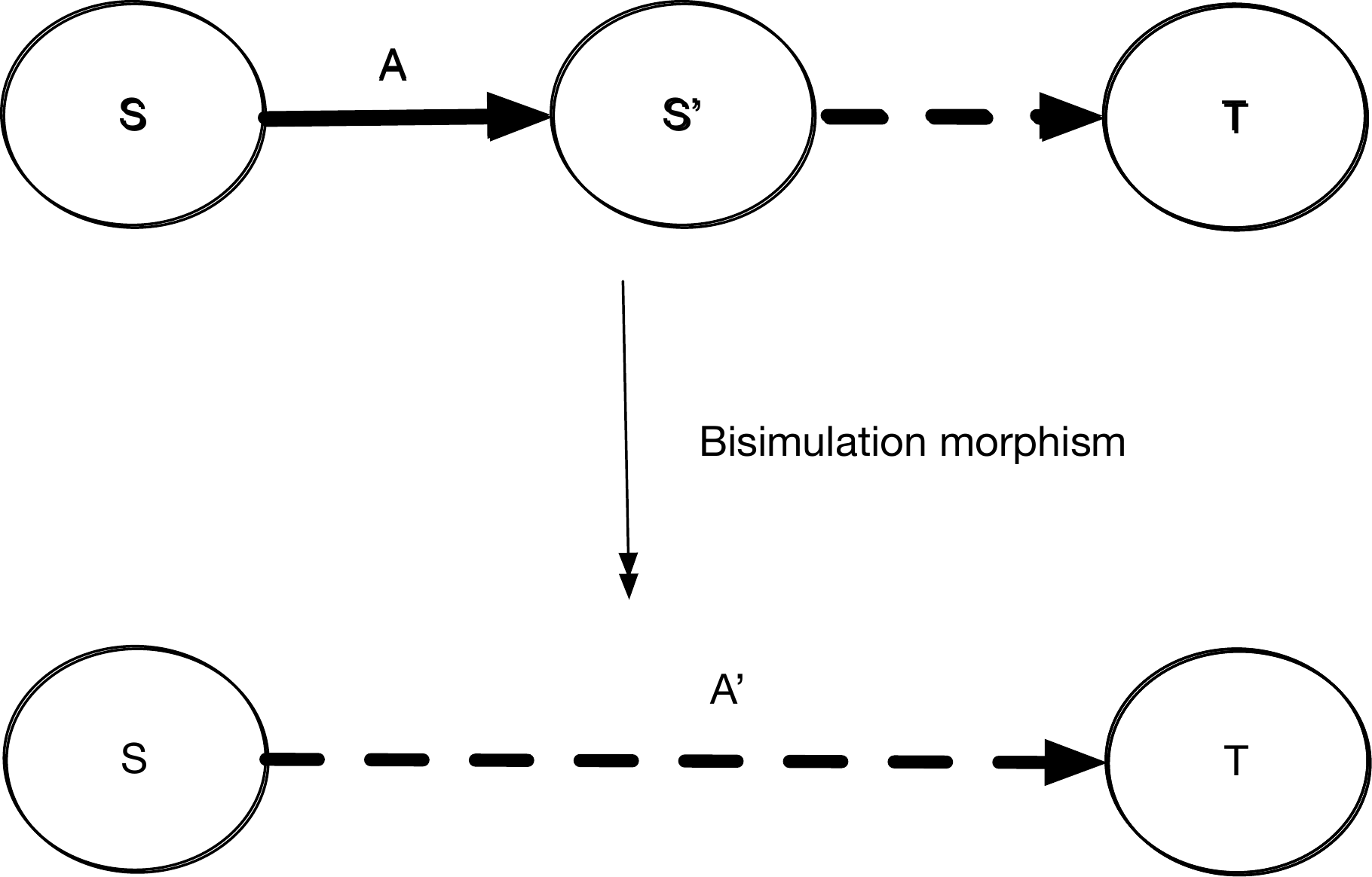}
\end{minipage}
\end{center}
\caption{A simple example of a bisimulation morphism between labeled transition systems \citep{DBLP:conf/lics/JoyalNW93}. Our work generalizes this construction to UDM models, replacing states by information fields.}
\label{bisim}
\end{figure}

In a UDM category, the morphisms between decision objects are represented using the concept of {\em open maps}, as proposed in \citep{DBLP:conf/lics/JoyalNW93}. This framework is based on defining a model of computation as a category.  Figure~\ref{bisim} illustrates a simple example of the concept of bisimulation in the category of labeled transition systems, which can be seen as a deterministic MDP \citep{DBLP:conf/lics/JoyalNW93}. A related notion has been proposed for probabilistic bisimulation \citep{DBLP:journals/iandc/LarsenS91}.  The use of category theory to provide an algebraic characterization of machine models has a long and distinguished history \citep{DBLP:conf/category/ArbibM74}, which  been studied at length in a number of different subfields of computer science.  One fundamental notion is {\em bisimulation} between machines or processes using {\em open maps} in categories \citep{DBLP:conf/lics/JoyalNW93,DBLP:journals/iandc/JoyalNW96}. This definition can be seen as a generalization of the simpler bisimulation relationship that exists for the category of labeled transition systems \citep{DBLP:conf/lics/JoyalNW93}, which are specified as a relation of tuples $(s, a, s')$, which  indicates a transition from state $s$ to state $s'$, where $s, s' \in S$, and $a \in A$.  Given a collection of labeled transition systems, each of which is represented as an object, morphisms are defined from one object to another that preserve the dynamics under the labeling function. For example, a surjective function $f: S \rightarrow S'$ maps states in object $X$ to corresponding states in $Y$, where the labels are mapped as well, with the proviso that some transitions in $X$ may be hidden in $Y$ (i.e., cause no transition). If a morphism exists between objects $X$ and $Y$, then $Y$ is said to be a bisimulation of $X$.

Let ${\cal M}$ denote a model of computation, where a morphism $m: X \rightarrow Y$ is to intuitively viewed as a simulation of $X$ in $Y$. Within ${\cal M}$, we choose a subcategory of ``observation objects" and ``observation extension" morphisms between them. We can denote this cateogry of observations by ${\cal P}$. Given an observation object $P \in {\cal P}$, and a model $X \in {\cal M}$, $P$ is said to be an {\em observable behavior} of $X$ if there is a morphism $o: P \rightarrow X$ in ${\cal M}$. We define morphisms $m: X \rightarrow Y$ that have the property that whenever an observable behavior of $X$ can be extended via $f$ in $Y$, that extension can be matched by an extension of the observable behavior in $X$. 

\begin{definition}\citep{DBLP:conf/lics/JoyalNW93}
\label{popen}
A morphism $m: X \rightarrow Y$ in a model of computation ${\cal M}$ is said to be ${\cal P}$-open if whenever $f: O_1 \rightarrow O_2$ in ${\cal P}$, $p: O_2 \rightarrow Y$ in ${\cal M}$, and $q: O_2 \rightarrow Y$ in ${\cal M}$, the below  diagram commutes, that is, $m \ p = q \ f$. 
\begin{center}
\begin{tikzcd}
  O_1 \arrow{d}{f} \arrow{r}{p}
    & X \arrow[red]{d}{m} \\
  O_2 \arrow[red]{r}[blue]{q}
&Y \end{tikzcd}
\end{center} 
\end{definition}
This definition means that whenever such a ``square" in ${\cal M}$ commutes, the path $f \ p$ in $Y$ can be extended via $m$ to a path $q$ in $Y$, there is a ``zig-zag" mediating morphism $p'$ such that the two triangles in the diagram below
\begin{center}
\begin{tikzcd}
  O_1 \arrow{d}{f} \arrow{r}{p}
    & X \arrow[red]{d}{m} \\
  O_2  \arrow[red]{ur}[blue]{p'} \arrow[red]{r}[blue]{q}
&Y \end{tikzcd}
\end{center} 
commute, namely $p = p' \ f$ and $q = m \ p'$. We now define the abstract definition of {\em bisimulation} as follows: 

\begin{definition}
Two models $X$ and $Y$ in ${\cal M}$ are said to be ${\cal P}$-bisimilar (in ${\cal M}$) if there exists a span of open maps from a common object $Z$: 
\begin{center}
 \begin{tikzcd}[column sep=small]
& Z \arrow{dl}{m} \arrow{dr}{m'} & \\
  X  &                         & Y
\end{tikzcd}
\end{center} 
\end{definition}

Note that if the category ${\cal M}$ has pullbacks (see Figure~\ref{univpr} right), then the $\sim_{{\cal P}}$ is an equivalence relation, which induces a quotient mapping. Furthermore, pullbacks of open map bisimulation mappings are themselves bisimulation mappings. Many of the bisimulation mappings studied for MDPs and PSRs are special cases of the more general formalism above. 

\subsection{Bisimulation in UDMs}

We introduce the concept of bisimulation morphisms between UDM objects, which builds on a longstanding theme in computer science on using category theory to understand machine behavior \citep{DBLP:conf/category/ArbibM74}. One fundamental notion is {\em bisimulation} between machines or processes using {\em open maps} in categories \citep{DBLP:conf/lics/JoyalNW93,DBLP:journals/iandc/JoyalNW96}. 

\begin{definition}
\label{bisim-defn}
The {\bf bisimulation} relationship between two UDM objects $M = \langle A, (\Omega, {\cal B}, (U_\alpha, {\cal F}_\alpha, {\cal I}_\alpha)_{\alpha \in A}\rangle$ and $M' = \langle A', (\Omega', {\cal B}', (U'_\alpha, {\cal F}'_\alpha, {\cal I}'_\alpha)_{\alpha \in A'}\rangle$,  denoted as $M \twoheadrightarrow M'$, is defined as  is defined by a tuple of surjections as follows: 
\begin{itemize}
    \item A surjection $f: A \twoheadrightarrow A'$ that maps  elements in $A$ to corresponding elements in $A'$. As $f$ is surjective, it induces an equivalence class in $A$ such that $x \sim y, x, y \in A$ if and only if $f(x) = f(y)$. 
    \item A surjection $g: H \twoheadrightarrow H'$, where $H = \Omega \times \prod_{\alpha \in A} U_\alpha$, with the product $\sigma$-algebra ${\cal B} \times \prod_{\alpha \in A} {\cal F}_\alpha$, and $H' = \Omega' \times \prod_{\alpha \in A'} U'_\alpha$, with the corresponding $\sigma$-algebra ${\cal B}' \times \prod_{\alpha \in A'} {\cal F}'_\alpha$. 
\end{itemize}
\end{definition}

This definition can be seen as a generalization of the simpler bisimulation relationship that exists for the category of labeled transition systems \citep{DBLP:conf/lics/JoyalNW93}, which are specified as a relation of tuples $(s, a, s')$, which  indicates a transition from state $s$ to state $s'$, where $s, s' \in S$, and $a \in A$.  Given a collection of labeled transition systems, each of which is represented as an object, morphisms are defined from one object to another that preserve the dynamics under the labeling function. For example, a surjective function $f: S \rightarrow S'$ maps states in object $X$ to corresponding states in $Y$, where the labels are mapped as well, with the proviso that some transitions in $X$ may be hidden in $Y$ (i.e., cause no transition). If a morphism exists between objects $X$ and $Y$, then $Y$ is said to be a bisimulation of $X$. Definition~\ref{bisim} can be seen as the generalization of the bisimulation relationship in labeled transition systems, MDPs, and related models like PSRs,  to intrinsic models. The state-dependent action recoding in the MDP homomorphism definition is captured by the equivalent surjection $g$ that maps the product space $H$ with its associated $\sigma$-algebra to $H'$ with its corresponding $\sigma$-algebra. 

We can specialize the definition in a number of ways, depending on the exact form chosen for the surjection $g$ between product spaces $H$ and $H'$. Since the surjection $f$ maps decision makers into equivalence classes, each decision maker in model $M'$ corresponds to an equivalence class of decision makers in model $M$. Thus, we need to collapse their corresponding information fields. We can define the information field of an equivalence class of agents $[\alpha]_f$, meaning all $\beta$ such that $f(\beta) = f(\alpha)$, by recalling that an information field is a subfield of the product field, and as it is a lattice, we can use the join operation, as defined below: 

\begin{definition}
The {\bf quotient information field} of a collection of agents $[\alpha]_f$ is defined as the join of the information fields of each agent: 
\begin{equation}
\label{qif}
{\cal I}_{[\alpha]} = \bigvee_{\beta \in [\alpha]_f} {\cal I}_\alpha
\end{equation}
\end{definition}

\subsection{Observation Objects in UDM}

We now briefly discuss observation objects in a UDM. Observation objects, as mentioned above, represent observable trace behavior of a decision object. We first define observation functions that underlie information fields.

\begin{definition}
For a UDM object $M = \langle A, (\Omega, {\cal B}, (U_\alpha, {\cal F}_\alpha, {\cal I}_\alpha)_{\alpha \in A} \rangle$ over a finite $\sigma$-algebra, the observations $Z_1, \ldots, Z_k$ taking values in a measurable space $(Z_i, {\cal Z}_i)$, where $Z_i = \eta_i(\omega, U_1, \ldots, U_{|A|})$ is an {\bf observation generation map} function  such that $\sigma(Z_{1:k})$ is the smallest $\sigma$-algebra contained in ${\cal B} \otimes \prod_\alpha {\cal F}_\alpha$ with respect to which the observation maps $\eta_i$ are measurable functions. We say the observations $Z_1, \ldots, Z_k$ generate the information field ${\cal I}_\alpha$ if 
\begin{equation}
    \sigma(Z_1, \ldots, Z_k) = {\cal I}_\alpha
\end{equation}
\end{definition}

We can then define an observation object associated with a UDM decision object as one equipped with an observation generation map that can generate the various information fields in the decision object. 

\begin{definition}
A UDM {\bf observation object} $O = \langle A, (\Omega, {\cal B}, (U_\alpha, {\cal F}_\alpha, \eta_\alpha)_\alpha \rangle$ is such that each information field ${\cal I}_\alpha$ can be generated from the associated observation generation map $\eta_\alpha$. 
\end{definition}

We can define an observation morphism between an observation object $O$ and a decision object $M$ to be one such that $O$ represents an observable behavior of $M$, and extend the notion of ${\cal P}$-open morphisms from Definition~\ref{popen} above. 

\subsection{Example: Network Economics} 

\begin{figure}[h]
\begin{center} \hfill
\begin{minipage}{0.75\textwidth}
\includegraphics[scale=0.4]{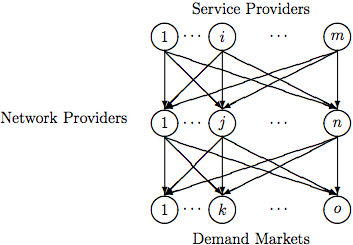}
\end{minipage} 
\end{center}
\caption{A multiplayer game network economic model \citep{nagurney:soi} as an example of a UDM. Each decision object in this UDM is a network economics model, where the top tier of producer agents is interested in selling merchandise (digital content, manufactured goods) to a set of demand market agents, but needs the cooperation of transport agents to deliver the merchandise. All the players in this network compete for the best price and quality.} 
\label{soi}
\end{figure}

To illustrate the general UDM framework, we now give an example of a multiplayer producer consumer game from network economics. Consider the network economic model in Figure~\ref{soi}. The set of elements in this decision object can be represented as $(A, (\Omega, {\cal B}, P), U_\alpha, {\cal F}_\alpha, {\cal I}_\alpha)_{\alpha \in A}$, where $A$ is defined by the set of vertices in this graph representing the decision makers. For example, service provider $i$ chooses its actions from the set $U_i$, which can be defined as $\cup_{j,k} Q_{ijk}$. ${\cal F}_i$ is the associated measurable space associated with $U_i$. ${\cal I}_i$ represents the information field of service provider $i$, namely its visibility into the decisions made by other entities in the network at the current or past time steps. 

 Network economics \citep{nagurney:vibook}is the study of a rich class of equilibrium problems that occur in the real world, from traffic management to supply chains and two-sided online marketplaces. Consider a cloud based network economics model comprises of three tiers of agents: producer agents, who want to sell their goods, transport agents who ship merchandise from producers, and demand market agents interested in purchasing the products or services. The model applies both to electronic goods, such as video streaming, as well as physical goods, such as face masks and other PPEs.  he model assumes $m$ service providers, $n$ network providers, and $o$ demand markets.  Each firm's utility function is defined in terms of the nonnegative service quantity (Q), quality (q), and price ($\pi$) delivered from service provider $i$ by network provider $j$ to consumer $k$.  Production costs, demand functions, delivery costs, and delivery opportunity costs are designated by $f$, $\rho$, $c$, and $oc$ respectively.  Service provider $i$ attempts to maximize its utility function $U_i^1(Q,q^*,\pi^*)$ by adjusting $Q_{ijk}$.  Likewise, network provider $j$ attempts to maximize its utility function $U_j^2(Q^*,q,\pi)$ by adjusting $q_{ijk}$ and $\pi_{ijk}$.

\begin{subequations}
\begin{align}
\label{U1}
U_i^1(Q,q^*,\pi^*) &= \sum_{j=1}^n \sum_{k=1}^o \hat{\rho}_{ijk}(Q,q^*)Q_{ijk} - \hat{f}_i(Q)
- \sum_{j=1}^n \sum_{k=1}^o \pi^*_{ijk}Q_{ijk}, \hspace{0.2cm} Q_{ijk} \ge 0 \nonumber
\end{align}
\begin{align}
U_j^2(Q^*,q,\pi) = &\sum_{i=1}^m \sum_{k=1}^o \pi_{ijk}Q^*_{ijk}
- \sum_{i=1}^m \sum_{k=1}^o (c_{ijk}(Q^*,q) + oc_{ijk}(\pi_{ijk})), \nonumber 
q_{ijk}, \pi_{ijk} \ge 0 \nonumber
\end{align}
\end{subequations}

As a second example, consider a  take a toy cloud computing network that is comprised of three elements $A = \{e_1, e_2, h \}$, where $e_i$ is an edge node, and $h$ is a hub node. Let us assume the decision space for the edge nodes $U_1 = U_2 = \{s, r \}$, representing ``send" and ``receive" modes, and for the hub $U_h = \{c, t \}$ representing ``collect" and ``transmit" modes. Let us also define the states of nature $\Omega = \{+, - \}$, indicating whether the environment is a ``safe" or ``unsafe" mode for information transmission or collection. Let us assume that the $\sigma$-algebras for both edge and hub devices is defined by the discrete topology given as ${\cal F}_{e_i} = \{\emptyset, \{ s \}, \{r \}, \{s, r \} \}, {\cal F}_h = \{ \emptyset, \{c \}, \{r \}, \{c, r \} \}$. The $\sigma$-algebra for states of nature is given as ${\cal B} = \{\emptyset , \{+ \}, \{ - \}, \{+, - \} \}$. The product decision space is given as $H = \Omega \times \prod_\alpha U_\alpha, \alpha \in \{e_1, e_2, h \}$, the product $\sigma$-algebra is defined as ${\cal F}_A = {\cal B} \times {\cal F}_{e_1} \times {\cal F}_{e_2} \times {\cal F}_h$. 

\subsection{Causality and Solvability of UDM objects} 

Each decision maker $\alpha$  in a UDM object  has associated with it a control law or policy $\pi_\alpha: H \rightarrow U_\alpha$, which is measurable from its information field ${\cal I}_\alpha$ to ${\cal F}_\alpha$, the measurable space associated with its decision space $U_\alpha$. Essentially, this means that any pre-image of $\pi^{-1}_\alpha(E)$, for any measurable subset $E \subset {\cal F}_\alpha$, is also  measurable on its information field, that is $\pi^{-1}_\alpha(E) \subset {\cal I}_\alpha$. For any subsystem in the cloud computing network, the overall policy space $\pi_B = \prod_{\alpha \in B} \pi_alpha$ is given by the product space of all individual control laws. 

\begin{definition} 
\label{solvable}
A UDM object $\langle A, (\Omega, {\cal B}, P), (U_\alpha, {\cal F}_\alpha, {\cal I}_\alpha)_{\alpha \in A} \rangle$ is said to be {\bf solvable} if for every state of nature $\omega \in \Omega$, and every control law $\pi \in \Pi_A$, the set of simultaneous equations given below has one and only one solution $u \in U$. 
\begin{equation}
    u_\alpha = \pi_\alpha(h) \equiv \pi_\alpha(\omega, u)
\end{equation}
Here, $\pi_\alpha$ can be viewed as a projection from the joint decision $h$ taken by the entire ensemble of decision makers in the intrinsic model. A UDM category ${\cal C}_{\mbox{UDM}}$ is solvable if every object in it is solvable. 
\end{definition} 

Intuitively, the solvability criterion states that a UDM object represents a solvable decision problem if each agent in the object can successfully compute its response, given access to its information field, and that its response is uniquely determined for every state of nature. It is easy to construct unsolvable decision objects. Consider a simple network with two elements $\alpha$ and $\beta$, each of whose information fields includes the measurable space of the other. In this case, neither element can compute its function without knowing the other's response, hence both are waiting for the other to compute their response, and a deadlock ensures. 

Given the notion of solvability above, we can now define solution objects in a UDM. 

\begin{definition} 
\label{solution}
A UDM {\bf solution object} $\langle A, (\Omega, {\cal B}, P), (U_\alpha, \pi_\alpha, {\cal F}_\alpha, {\cal I}_\alpha)_{\alpha \in A} \rangle$ is defined as one for which for every state of nature $\omega \in \Omega$, the control law $\pi_\alpha$ uniquely defines a fixed point solution $u_\alpha = \pi_\alpha(h) \equiv \pi_\alpha(\omega, u)$ to the associated decision object. 
\end{definition} 

We can straightforwardly define morphisms between solution objects and decision objects. 
To understand the causality condition, it is crucial to organize the decision makers into a partial order, such that for every total ordering that can be constructed from the partial ordering, the agents can successfully compute their functions based on the computations of agents that preceded them in the ordering. 

\begin{definition} 
\label{causal}
A UDM object  $\langle A, (\Omega, {\cal B}, P), (U_\alpha, {\cal F}_\alpha, {\cal I}_\alpha) \rangle$ is said to be {\bf causal} if there exists at least one function $\phi: H \rightarrow S$, where $S$ is the set of total orderings of computing elements in $A$, satisfying the property that for any partial stage of the computation $1 \leq k \leq n$, and any ordered set $(\alpha_1, \ldots, \alpha_k)$ of distinct elements from $A$, the set $E \subset H$ on which $\phi(h)$ begins with the same ordering $(\alpha_1, \ldots, \alpha_k)$ satisfies the following causality condition: \begin{equation} 
\label{causality} 
\forall F \in {\cal F}_{\alpha_k},\ \ E \cap F \in {\cal F}(\{\alpha_1, \ldots, \alpha_{k-1} \})
\end{equation} 
\end{definition} 
In other words, if at every step of the process, the $k^{\mbox{th}}$ decision making element $\alpha_k$ can successfully compute its response based on the information fields of the past $k-1$ elements, the system is then considered causal. Interestingly, it has been shown (see \citep{DBLP:journals/corr/heymann}) that the causality condition as stated above generalizes the notion of causality in Pearl's structural causal models \citep{pearl-book}. 

\section{UDMs for Causal Inference and Stochastic Control} 

We now show the general framework of UDMs transcends multiple decision making regimes, by illustrating how they can form the basis for "universal" decision making in two special cases: linear total ordering, which gives rise to stochastic control, and partial ordering, which gives rise to causal inference. 

\subsection{UDMs in  RL and Stochastic Control} 

\citet{DBLP:journals/mst/Witsenhausen73} himself showed the importance of information fields in stochastic control, in particular developing a canonical model of stochastic control \citep{DBLP:journals/mst/Witsenhausen73}. We summarize a more recent extension from \citep{DBLP:journals/tac/NayyarT19} that shows how to define common knowledge using information fields, an interesting contrast to the notion of common knowledge defined above in game theory using Aumann's framework. We define the abbreviation $U_{1:T} = (U_1, \ldots, U_T)$ for the product decision space, and similarly ${\cal F}_{1:T} = {\cal F}_1 \times \ldots \times {\cal F}_T$ for the product $\sigma$-algebra. 

\begin{definition} \citep{DBLP:journals/tac/NayyarT19}
\label{im-control}
Given the probability model $(\Omega, {\cal B}, P)$ for the random states of nature $\omega \in \Omega$, the measurable decision spaces $(U_t, {\cal F}_t), t = 1, \ldots, T$, the information field $\sigma$-algebras ${\cal I}_t \subset {\cal B} \times {\cal F}_1 \ldots {\cal F}_T$, and the cost function $c: (\Omega \times U_{1:T} \times, {\cal B} \times {\cal F}_{1:T} \rightarrow (\mathbb{R}, \mathbb{B})$, find a (generally non-stationary) policy $\pi = (\pi_1, \ldots, \pi_T)$, with each policy at time $t$ defined as the mapping $g_t: (\Omega \times U_{1:T}, {\cal I}_t \rightarrow (U_t, {\cal F}_t)$, that minimizes the cost function $\inf_\pi E[c(\omega, U_1, \ldots, U_T)]$ exactly, or to within $\epsilon$. 

\end{definition}

Note that the above definition of stochastic  decision making is just a special case of the UDM model  in Definition~\ref{udm-defn}. In particular, the agents in stochastic control are labeled $1, \ldots, T$, their temporal ordering is fixed a priori, and each agent's information field is generally defined over the entire horizon $(1, \ldots, T)$. To define the special case of finite horizon {\em sequential} stochastic control, we must impose further conditions on the information fields available at each instant of time $t \in (1, \ldots, T)$. 

\begin{definition}
An information structure in the stochastic control model in Definition~\ref{im-control} is {\bf sequential} if there exists a permutation $p: \{1, \ldots, T\} \rightarrow \{1, \ldots, T \}$ such that for $t = 1, \ldots T$, the information field ${\cal I}_t \subset {\cal B} \times {\cal F}_{p(1)} \times {\cal F}_{p(2)}, \ldots, {\cal F}_{p(t-1)} \times \{\emptyset, {\cal F}_{p(t)} \} \times \ldots \times \{ \emptyset, {\cal F}_{p(T)} \}$. 
\end{definition}

In terms of the terminology we have introduced earlier, note that the information field $I_t$ at time $t$ is a cylindrical extension from the field over $1, \ldots, t-1$ to all of $1, \ldots, T$. Note that for the sequential case, the permutation ordering $p$ is fixed a priori, and does not vary over the different states of nature $\omega \in \Omega$. 

We now define the notion of {\em common knowledge} in information fields based on the definition in \citep{DBLP:journals/tac/NayyarT19}. Recall that the information field ${\cal I}_t \subset {\cal F} \times {\cal F}_1 \times \ldots {\cal F}_{t-1} \times \{\emptyset, U_t \} \times \ldots \{ \emptyset, U_T \}$. 

\begin{definition} \citep{DBLP:journals/tac/NayyarT19}
The {\bf common knowledge} for the $t^{\mbox{th}}$ decision maker in a sequential intrinsic model is defined as 
\begin{equation}
    {\cal C}_t = \bigcap_{s = t}^T {\cal I}_s
\end{equation}

That is, the common knowledge ${\cal C}_t$ is defined as the intersection of all information fields from time $t$ till the end of the decision process. 
\end{definition}

Some simple properties of common knowledge can be readily shown: 

\begin{itemize}
    \item Coarsening property: ${\cal C}_t \subset {\cal I}_t$: immediate from definition. 
    
    \item Nestedness property: ${\cal C}_t \subset {\cal C}_{t+1} $: immediate from definition. 
    
    \item Common observations: There exist observations $Z_1, \ldots, Z_T$ with $Z_t$ taking values in a finite measurable space $(Z_t, 2^{Z_t})$, and $Z_t = \eta_t(\omega, U_1, \ldots, U_{t-1})$ such that $\sigma(Z_{1:t}) = {\cal C}_t$: for a detailed proof, see \citep{DBLP:journals/tac/NayyarT19}). The basic idea exploits the fact that finite $\sigma$-algebras can be generated from partitions. 
\end{itemize}

\subsection{The Category of Causal UDMs} 

\label{causal-if}

In the above, we assume that temporal ordering is given  a priori as a total ordering $(1, \ldots, T)$, and in the sequential case, each decision maker's information field is a subset of the product decision and information fields of all agents that have acted prior to it. We now generalize from the requirement of imposing a strict linear ordering, and consider more general partially ordered temporal structures. This relaxation from linear to partial ordering allows us to formalize a particular case of the intrinsic model that in fact exactly corresponds to causal inference, as shown recently in \cite{DBLP:journals/corr/heymann}.  We briefly review how information fields can formalize causal inference, referring the interested reader to \citep{DBLP:journals/corr/heymann} for additional details. Consider a simple causal model shown below, where variable $A$ is a ``common" cause of variables $B$ and $C$. In a structural causal model \citep{pearl-book}, we consider the universe of variables $\{A, B, C \}$ to be subdivided into ``exogenous" variables $U$  with no parents in the model, below $U = \{A \}$, and ``endogenous" variables $V = \{B, C \}$ whose parents include exogenous and endogenous variables. 

\begin{center}
 \begin{tikzcd}[column sep=small]
& A \arrow[dl] \arrow[dr] & \\
  B \arrow{rr} &                         & C
\end{tikzcd}
\end{center} 

Let us illustrate how information fields can be used to represent such structural causal models. Let the three variables above all be binary, so each variable can be viewed as a decision maker whose decision space $U_A = U_B = U_C = \{0, 1\}$. Let the associated $\sigma$-algebras be defined as by the discrete topology ${\cal F}_A = {\cal F}_B = {\cal F}_C = \{\emptyset, \{0 \}, \{1 \}, \{0, 1\} \}$. Let the states of nature be defined as $\Omega = \{0, 1\}^3$, with the associated Borel topology ${\cal B} = 2^\Omega$. We can think of $\Omega = \Omega_A \times \Omega_B \times \Omega_C$, and ${\cal B} = {\cal B}_A \times {\cal B}_A \times {\cal B}_C$. To specify the causal DAG model fully, we need to specify the conditional probability distributions, which we can do using information fields for each variable. 

Consider the exogenous variable $A$. Since it has no parent in the model, its value depends only on the measure of uncertainty from the external environment, hence we can write its information field ${\cal I}_A \subset {\cal B}_A \times\{\emptyset, \Omega_B \} \times \{\emptyset, \Omega_C \} \times \{\emptyset, U_A \}$. Note that a variable in a structural causal model cannot be ``self-aware", that is, its value cannot depend on its own value! Hence, the condition $\{\emptyset, U_A \}$ is imposed. On the other hand, the information field for variable $C$ depends on the values of the other two variables, so its information field can be written as ${\cal I}_C \subset \{\emptyset, \Omega_A \} \times \{\emptyset, \Omega_B \} \times {\cal F}_A \times {\cal B}_C \times {\cal F}_B \times \{\emptyset, U_C \}$. That is, the value taken by $C$ depends on the values taken by $A$ and $B$ and its own uncertainty. 

\begin{definition}
A {\bf causal UDM} is defined as one where each object ${\cal M} = (U_\alpha, {\cal F}_\alpha, {\cal I}_\alpha, (\Omega, {\cal B}, P))$, where $\alpha \in X$, a finite space of variables. $U_\alpha$ is a non-empty set that defines the range of values that variable $\alpha$ can take. ${\cal F}_\alpha$ is a $\sigma$-algebra of measurable sets for variable $\alpha$. The triple $(\Omega, {\cal B} , P)$ is a probability space, where ${\cal B}$ is a $\sigma$-algebra of measurable subsets of sample space $\Omega$. The {\bf information field} ${\cal I}_\alpha \subset {\cal F}$ represents the ``receptive field" of an element $\alpha \in X$, namely the set of other elements $\beta \in X$ whose values $\alpha$ must consult in determining its own value. We impose the restriction that the information field ${\cal I}_\alpha$ respect the Alexandroff topology on $X$, so that ${\cal I}_\alpha \subset {\cal F}(U_\alpha)$, where $U_\alpha$ is the minimal basic open set associated with element $\alpha \in X$.
\end{definition}

Following structural causal models \citep{pearl-book}, we can decompose the elements of a causal UDM object into disjoint subsets $X = U \sqcup V$, where $U$ represents ``exogenous" variables that have no parents, namely $\alpha$ is exogenous precisely when ${\cal I}_\alpha \subset {\cal F}(\emptyset)$, and $V$ are ``endogenous" variables whose values are defined by measurable functions over exogenous and endogenous variables. Note that the probability space can be defined over the ``exogenous" variables $\alpha \in U$, in which case it is convenient to attach a local probability space $(\Omega_\alpha, {\cal B}_\alpha, P)$ to each exogenous variable, where ${\cal B}_\alpha \subset {\cal B}$. We define conditional independence with respect to the induced information fields over the open sets of the Alexandroff space. 

\begin{definition}
\label{ci-top}
Given the induced probability space over information fields in a causal UDM object, a {\em stochastic basis} is a sequence of information fields ${\cal G} = {\cal I}_1, \ldots, {\cal I}_n$ such that for $1 \leq i \leq n-1, {\cal I}_i \subset {\cal I}_{i+1}$, and $\cup_{1=1}^n {\cal I}_i = {\cal F}$. Two such sequences ${\cal G}_1$ and ${\cal G}_2$ are {\bf conditionally independent} given the base $\sigma$-algebra ${\cal F}$, if for all subsets $A \in {\cal G}_1$, $B \in {\cal G}_2$, it follows that
$P(A \ B | {\cal F} )= P(A | {\cal F}) P(B | {\cal F})$. 
\end{definition}

\begin{definition}
The {\bf decision field} $U = \prod_{\alpha \in X} U_\alpha$ defines the space of all possible values of the variables in a causal UDM object, where the cartesian product is interpreted as a map $u: X \rightarrow \cup_{\alpha \in X} U_\alpha$ such that $u(\alpha) \equiv u_\alpha \in U_\alpha$.  
\end{definition}

\begin{definition}
For any subset of elements $B \in X$, let $P_B$ denote the {\em projection} of the product $\prod_\alpha U_\alpha$ upon the product $\prod_{\beta \in B} U_\beta$, that is $P_B(u)$ is simply the restriction of $u$ to the domain $B$. 
\end{definition}

\begin{definition}
The product $\sigma$-algebra is defined as $\prod_{\alpha \in B} {\cal F}_\alpha$ over $\prod_{\alpha \in B} U_\alpha$, where ${\cal F}(B)$ is the smallest sigma-field such that $P_B$ is measurable. Note that if $B_1 \subset B_2$, then ${\cal F}_{B_1} \subset {\cal F}_{B_2}$. The finest sigma-field ${\cal F}(X) = \prod_{\alpha \in X} {\cal F}_\alpha$.  
\end{definition}

\begin{definition}
A causal UDM object  ${\cal M}$ is {\bf causally faithful} with respect to the probability distribution $P$ over ${\cal M}$ if every conditional independence in the topology, as defined in Definition~\ref{ci-top}, is satisfied by the distribution $P$, and vice-versa, every conditional independence property of the $P$ is satisfied by the topology. 
\end{definition}

We can now formally define what it means to ``solve" a causal UDM object ${\cal M}$. We impose the requirement that each variable $\alpha \in X$ must compute its value using a function measurable on its own information field. 

\begin{definition}
Let the policy function $f_\alpha$ of each element $\alpha \in X$ be constrained so that $f_\alpha: U \times \Omega \rightarrow U_\alpha$ is measurable on the product $\sigma$-algebra ${\cal I}_\alpha \times {\cal B}_\alpha$, namely $f_\alpha^{-1}({\cal F}_\alpha) \subset {\cal I}_\alpha \times {\cal B_\alpha}$. 
\end{definition}

\begin{definition}
\label{property-sm1}
The causal UDM object ${\cal M} = (X, U_\alpha, {\cal F}_\alpha, {\cal I}_\alpha)$ is {\bf measurably solvable} if for every $\omega \in \Omega$, the closed loop equations $P_\alpha(u) = f_\alpha(u, \omega)$ have a unique solution for all $\alpha \in X$, where for a fixed $\omega \in \Omega$, the induced map ${\cal M}^\gamma: \Omega \rightarrow U$ is a measurable function from the measurable space $(\Omega, {\cal B})$ into $(U, {\cal F})$. 
\end{definition}

\begin{definition}
\label{property-sm2}
The causal UDM object ${\cal M} = (X, U_\alpha, {\cal F}_\alpha, {\cal I}_\alpha)$ is {\bf stable} if for every $\omega \in \Omega$, the closed loop equations $P_\alpha(u) = f_\alpha(u, \omega)$ are solvable by a fixed constant ordering $\Xi$ that does not depend on $\omega \in \Omega$. 
\end{definition}

Measurably solvable causal UDM objects generalize the corresponding property in a structural causal model $(U,V,F,P)$, which states that for any fixed probability distribution $P$ defined over the exogenous variables $U$, each function $f_i$ computes the value of variable $x_i \in V$, given the value of its parents $Pa(x_i)$ uniquely as a function of $u \in U$. This allows defining the induced distribution $P_u(V)$ over exogenous variables in a unique functional manner depending on some particular instantiation of the random exogenous variables $U$. Stable models are those where the ordering of variables is fixed.  We now extend the notion of {\em recursive} causal models in DAGs \citep{pearl-book} to finite topological spaces. 

\begin{definition}
The causal UDM object ${\cal M} = (X, U_\alpha, {\cal F}_\alpha, {\cal I}_\alpha)$ is a {\bf recursively causal} model if there exists an ordering function $\psi: X \rightarrow \Xi_n$, where $\Xi_n$ is the set of all injective (1-1) mappings of $(1, \ldots, n)$ to the set $X$, such that for any $ 1 \leq k \leq n$, the information field of variable $\alpha_k$ in the ordering $\Xi_n$ is contained in the joint information fields of the variables preceding it:
\begin{equation}
    {\cal I}_{\alpha_k} \subset {\cal F}(\alpha_1, \ldots, \alpha_{k-1})
\end{equation}
\end{definition}

In other words, the ordering $\psi$ essentially proves a filtration of the $\sigma$-algebras over the previous variables to make the causal UDM object solvable. Note this property generalizes the recursive property in DAG models. What recursively causal means in the above definition is that element $\alpha_k$ has the information needed to compute its value based on the values of the variables that preceded it in the ordering given by $\psi$, and crucially, this ordering need not be the same for every element $\omega \in \Omega$ in the sample space. That is, for some setting of the exogenous variables, it may very well turn out that the ordering changes. This variability is not the case in DAG models, where there is an assumption of a fixed ordering on the DAG induced by the partial ordering, which is independent of any randomness in the exogenous variables. Finally, we define causal interventions in finite topological spaces prior to describing algorithms for learning causal finite space models.

\begin{definition}
A {\bf causal intervention}  do$(\beta$=$u_\beta)$ in a causal UDM object ${\cal M} = (X, U_\alpha, {\cal F}_\alpha, {\cal I}_{\alpha})$ is defined as the subobject ${\cal M}_\beta$ whose information fields ${\cal I}_\alpha$ are exactly the same as in $M$ for all elements $\alpha \neq \beta$, and the information field of the intervened element $\beta$ is defined to be ${\cal I}_\beta \subset {\cal F}(\emptyset) \times {\cal B}_\beta$. Note that since the only measurable function on ${\cal F}(\emptyset)$ is the constant function, whose value depends on a random sample space element $\omega \in \Omega_\beta$, this generalizes the notion of causal intervention in DAGs, where an intervened node has all its incoming edges deleted. \footnote{Our definition of causal intervention differs from that proposed in causal information fields \citep{DBLP:journals/corr/heymann}, where additional intervention nodes were added to the model.} 
\end{definition}

\subsection{ UDMs based on Markov Decision Processes}

We now briefly describe the (sub) category of UDMs, where each object represents a (finite) Markov decision process (MDP) \citep{DBLP:books/wi/Puterman94}.  Recall that an MDP is defined by a tuple $\langle S, A, \Psi, P, R \rangle$, where $S$ is a discrete set of states, $A$ is the discrete set of actions, $\Psi \subset S \times A$ is the set of admissible state-action pairs, $P: \Psi \times S \rightarrow [0,1]$ is the transition probability function specifying the one-step dynamics of the model, where $P(s,a,s')$ is the transition probability of moving from state $s$ to state $s'$ in one step under action $a$, and $R: \Psi \rightarrow \mathbb{R}$ is the expected reward function, where $R(s,a)$ is the expected reward for executing action $a$ in state $s$. MDP homomorphisms can be viewed as a principled way of abstracting the state (action) set of an MDP into a ``simpler" MDP that nonetheless preserves some important properties, usually referred to as the stochastic substitution property (SSP). 

\begin{definition}
A UDM MDP homomorphism \citep{DBLP:conf/ijcai/RavindranB03} from object  $M = \langle S, A, \Psi, P, R \rangle$ to $M' = \langle S', A', \Psi', P', R' \rangle$, denoted $h: M \twoheadrightarrow M'$, is defined by a tuple of surjections $\langle f, \{g_s | s \in S \} \rangle$, where $f: S \twoheadrightarrow S', g_s: A_s \twoheadrightarrow A'_{f(s)}$, where $h((s,a)) = \langle f(s), g_s(a) \rangle$, for $s \in S$, such that the stochastic substitution property and reward respecting properties below are respected: 
\begin{eqnarray} 
\label{mdp-hom}
P'(f(s), g_s(a), f(s')) = \sum_{s" \in [s']_f} P(s, a, s") \\
R'(f(s), g_s(a)) = R(s, a)
\end{eqnarray} 
\end{definition}

Given this definition, the following result is straightforward to prove. 

\begin{theorem}
The UDM category ${\cal C}_{\mbox{MDP}}$ is defined as one where each object $c$ is defined by an MDP, and morphisms are given by MDP homomorphisms defined by Equation~\ref{mdp-hom}. 
\end{theorem}

{\bf Proof:} Note that the composition of two MDP homomorphisms $h: M_1 \rightarrow M_2$ and $h': M_2 \rightarrow M_3$ is once again an MDP homomorphism $h' \ h: M_1 \rightarrow M_3$. The identity homomorphism is easy to define, and MDP homomorphisms, being surjective mappings, obey associative properties. $\qed$

\subsection{UDM Category of Predictive State Representations} 

We now define the UDM (sub)category ${\cal C}_{\mbox{PSR}}$ of predictive state representations \citep{DBLP:journals/jmlr/ThonJ15}, based on the notion of homomorphism defined for PSRs proposed in  \citep{DBLP:conf/aaai/SoniS07}.  Recall that a PSR is (in the simplest case) a discrete controlled dynamical system, characterized by a finite set of actions $A$, and observations $O$. At each clock tick $t$, the agent takes an action $a_t$ and receives an observation $o_t \in O$. A {\em history} is defined as a sequence of actions and observations $h = a_1 o_1 \ldots a_k o_k$. A {\em test} is a possible sequence of future actions and observations $t = a_1 o_1 \ldots a_n o_n$. A test is successful if the observations $o_1 \ldots o_n$ are observed in that order, upon execution of actions $a_1 \ldots a_n$. The probability $P(t | h)$ is a prediction of that a test $t$ will succeed from history $h$. 

A state $\psi$ in a PSR is a vector of predictions of a suite of {\em core tests} $\{q_1, \ldots, q_k \}$. The prediction vector $\psi_h = \langle P(q_1 | h) \ldots P(q_k | h) \rangle$ is a sufficient statistic, in that it can be used to make predictions for any test. More precisely, for every test $t$, there is a $1 \times k$ projection vector $m_t$ such that $P(t | h) = \psi_h . m_t$ for all histories $h$. The entire predictive state of a PSR can be denoted $\Psi$. 

\begin{definition}
\label{psr-homo}
In the UDM category ${\cal C}_{\mbox{PSR}}$ defined by PSR objects, the morphism from object $\Psi$ to another  $\Psi'$ is defined by a tuple of surjections $\langle f, v_\psi(a)\rangle$, where $f: \Psi \rightarrow \Psi'$ and $v_\psi: A \rightarrow A'$ for all prediction vectors $\psi \in \Psi$ such that 
\begin{equation}
    P(\psi' | f(\psi), v_\psi(a)) = P(f^{-1}(\psi') | \psi, a) 
\end{equation}
for all $\psi' \in \Psi, \psi \in \Psi, a \in A$. 
\end{definition}

\begin{theorem}
The UDM category ${\cal C}_{\mbox{PSR}}$ is defined by making each object $c$ represent a PSR, where the morphisms between two PSRs $h: c \rightarrow d$ is defined by the PSR homomorphism defined in \citep{DBLP:conf/aaai/SoniS07}. 
\end{theorem}

 {\bf Proof:} Once again, given the homomorphism definition in Definition~\ref{psr-homo}, the UDM category ${\cal P}_{\mbox{PSR}}$ is easy to define, given the surjectivity of the associated mappings $f$ and $v_\psi$. $\qed$

\section{Topology associated with UDM objects} 
\label{im-topology} 

\begin{figure}[t]
\begin{center}
\begin{minipage}{0.5\textwidth}
\includegraphics[scale=0.4]{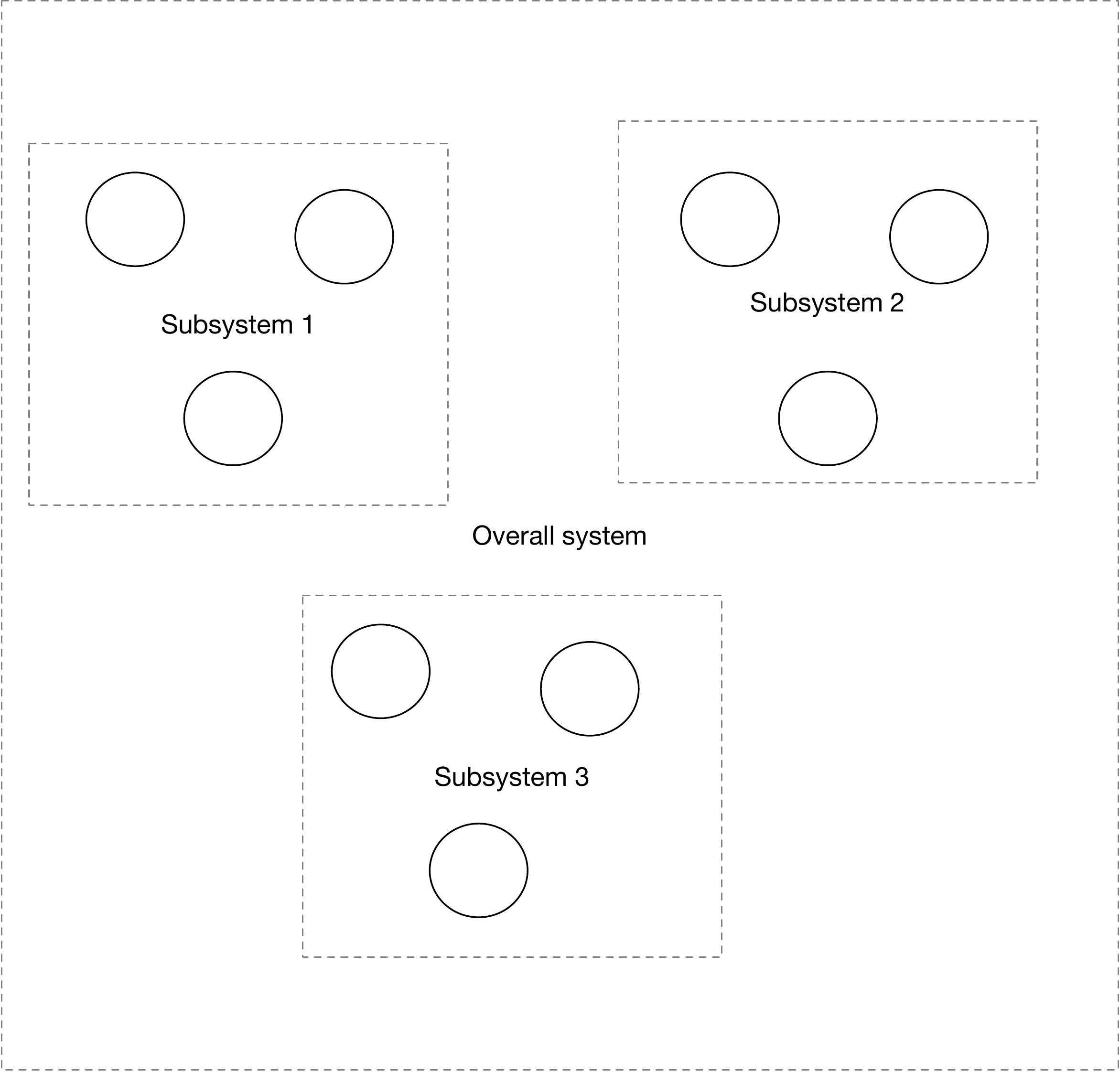}
\end{minipage}
\end{center}
\caption{Organization of a UDM object into subsystems is based on a finite space topology induced by the information field structure.}
\label{subsys}
\end{figure}

Figure~\ref{subsys} illustrates a simple way to decompose a UDM object into sub-objects. The information field structure induces a finite space topology that enables decomposing a complex UDM object into subobjects. We define the closure of an element $\alpha \in A$ as the set of elements on whom it depends for information. These closure sets will define a finite topology on the space of decision makers, enabling the decomposition of complex objects into more manageable pieces. The induced topology has a rich structure, and has many consequences for organizing computation. 

\begin{definition} 
\label{subsystem-defn}
A subset of decision makers $B \subset A$ in a UDM object form a {\bf subsystem} if the data requirements of members of the set only depend on the actions of nature, and the actions of the members of the set, and is independent of the actions of the non-members. More precisely, $B \subset A$ is a subsystem if for all $\alpha \in B, \ {\cal I}_\alpha \subset {\cal F}(B)$. If $B$ is a subsystem, the induced UDM object  $\langle B, (\Omega, {\cal B}, P), (U_\alpha, {\cal F}_\alpha, {\cal I}_{\alpha B})_{\alpha \in B} \rangle$ is also a valid UDM object by itself, where the induced information subfield ${\cal I}_{\alpha B}$ is the canonical projection of ${\cal I}_B$ upon $H_B$. 
\end{definition} 

\begin{definition} 
The {\bf closure} of a decision maker $\alpha \in A$ in a UDM object  $\langle A, (\Omega, {\cal B}, P), (U_\alpha, {\cal F}_\alpha, {\cal I}_\alpha)_{\alpha \in A} \rangle$ is the smallest subsystem containing $\alpha$, denoted by $\overline{\{\alpha \}}$. 
\end{definition}

\begin{definition}
The {\bf preorder} relationship between decision makers, denoted $\alpha \leftarrow \beta$ is defined by the containment between the closure sets, namely $\alpha \leftarrow \beta$ if and only if $\overline{\{ \alpha \}} \subset \overline \{ \beta \}$. 
\end{definition}

Note that the $\rightarrow$ relation defined above is a preorder because it is clearly reflexive and transitive. To explore more interesting special cases of this relationships, we need to introduce some additional notions from the topology of finite spaces. 

\begin{theorem}
The subsystems of a UDM object  $\langle A, (\Omega, {\cal B}, P), (U_\alpha, {\cal F}_\alpha, {\cal I}_\alpha)_{\alpha \in A} \rangle$ induce a finite space topology on the space $A$ of decision makers. 
\end{theorem}

{\bf Proof:} (adapted from \citep{witsenhausen:75}): Recall that in a finite space topology \citep{barmak}, the collection of subsets of $A$ termed ``open" sets are closed under arbitrary unions and intersections (it's worth pointing out that in the general case, topological spaces require finite intersections). As the complement of a open set is a closed set, the set of closed sets is also closed under intersection and union. Given two subsystems $S_1$ and $S_2$, if  element $\alpha \in S_1 \cup S_2$, then either ${\cal I}_\alpha \subset {\cal F}(S_1)$ or ${\cal I}_\alpha \subset {\cal F}(S_2)$. It follows that ${\cal I}_\alpha \subset {\cal F}(S_1) \cup {\cal F}(S_2) = {\cal F}(S_1 \cup S_2)$. The proof for closure under intersection is similar. $\qed$

Given this theorem, we can immediately bring to bear the powerful tools of algebraic topology \citep{barmak} of finite topological spaces, also called Alexandroff spaces \citep{alexandroff:1937}, to analyze the topological properties of UDM objects. Essentially, we are showing that UDMs form a subcategory in the category of all topological spaces (as each UDM object is a topological space in its own right). We briefly review some of the key properties that we will use below. 

\begin{definition}
\label{t0}
The {\em neighborhood} of an element $x$ in a finite space $X$ is a subset $V \subset X$ such that $x \in U$ for some open set $U \subset V$. 
\begin{itemize} 
\item $X$ is a  Kolmogorov (or $T_0$) finite space $X$ if each pair of points $x, y \in X$ is distinguishable in the space, namely for each $x, y \in X$, there is an open set $U \in \mathcal{U}$ such that $x \in U$ and $y \notin U$. Alternatively, if $x \in U$  if and only if $y \in U,  \ \forall U \in \mathcal{U}$ implies that $x = y$. 
\item $X$ is a $T_1$ finite space if element $x \in X$ defines a closed set $\{ x \}$. 
\item $X$ is a $T_2$ finite space or a {\em Hausdorff} space if any two points have distinct neighborhoods.
\end{itemize}
\end{definition}

\begin{lemma}
If $X$ is a $T_2$ space, then it is a $T_1$ space. If $X$ is a $T_1$ space, then it is a $T_0$ space.
\end{lemma}

The key concept that gives finite (Alexandroff) spaces its power is the definition of the minimal open basis. First, we introduce the concept of a basis in a topological space. 

\begin{definition}
A {\em basis} for the topological space $X$ is a collection $\mathcal{B}$ of subsets of $X$ such that 
\begin{itemize} 
\item For each $x \in X$, there is at least one $B \in \mathcal{B}$ such that $x \in B$.
\item If $x \in B' \cap B"$, where $B, B" \in \mathcal{B}$, then there is at least one $B \in \mathcal{B}$ such that $x \in B \subset B' \cap B"$. 
\end{itemize}
\end{definition}
The topology $\mathcal{U}$ {\em generated} by the basis $\mathcal{B}$ is the set of subsets $U$ such that for every $x \in U$, there is a $B \in \mathcal{B}$ such that $x \in B \subset U$. In other words, $U \in \mathcal{U}$ if and only if $U$ can be generated by taking unions of the sets in the basis $\mathcal{B}$. Now, we turn to giving the most important definition in Alexandroff spaces, namely the {\em unique minimal basis}. 

\begin{lemma}
Let $X$ be a finite Alexandroff space. For each $x \in X$, define the open set $U_x$ to be the intersection of all open sets that contain $x$. Define the relationship $\leq$ on $X$ by $x \leq y$ if $x \in U_y$, or equivalently, $U_x \subset U_y$ (where $x < y$ if the inclusion is strict). The open sets $U_x$ constitute a {\bf unique minimal  basis} $\mathcal{B}$ for $X$ in that if $\mathcal{C}$ is another basis for $X$, then $\mathcal{B} \subset \mathcal{C}$. Alternatively, define the closed sets $F_x = \{y \ | \ y \geq x \}$, which provide an equivalent characterization of finite Alexandroff spaces.\footnote{The minimal basic closed sets in a $T_0$ finite Alexandroff space correspond to the {\em ancestral sets} in a DAG graphical model.} 
\end{lemma}

Note that the relation $\leq$ defined above is a {\em preorder} because it is reflexive (clearly, $x \in U_x$) and transitive (if $x \in U_y$, and $y \in U_z$, then $x \in U_z$). However, in the special case where the finite space $X$ has a $T_0$ topology, then the relation $\leq$ becomes a partial ordering. 

\subsection{Classes of UDMs} 
\label{im-subsys}

We now describe a way to decompose UDM objects based on information fields. \cite{witsenhausen:75} defines the following $10$ classes of information structures, each of which leads to a distinct type of UDM object. This decomposition shows the importance of the topology induced on a UDM object based on information structures. 

\begin{enumerate} 

\item {\bf Monic:} A monic  UDM object has only decision maker $A = \{ \alpha \}$, and its information field ${\cal I}_\alpha \subset {\cal F}(\emptyset)$. In other words, the decision maker $\alpha$ only requires access to the state of nature, and does not obviously need information from any other decision maker, including itself! 

\item {\bf Team:} A team UDM object can be viewed as an independent set of decision makers, all of whom only need access to the state of nature, that is ${\cal I}_\alpha \subset  {\cal F}(\emptyset)$. 

\item {\bf Sequential: } A sequential UDM object is one where there exists a fixed ordering $\{\alpha_1, \ldots, \alpha_n \}$ of  decision makers from $A$ such that for any $1 \leq k \leq n$, it holds that ${\cal I}_{\alpha_k} \subset {\cal F}(\{ \alpha_1, \ldots, \alpha_{k-1} \}$. Sequential systems satisfy the causality condition with a constant ordering function $\phi$. 

\item {\bf Classical:} A UDM object is called classical if it is sequential, and furthermore, ${\cal I}_0 \in {\cal F}(\emptyset)$, ${\cal I}_{k-1} \subset {\cal I}_k$, for all $k  = 2 \ldots, n$. 

\item {\bf Strictly classical:} A UDM object is strictly classical if it is classical, and $[ {\cal F}_{\alpha_k} ] \subset {\cal I}_{k+1}$, where $[ {\cal F}_{\alpha_k}]$ is the cylindrical extension of ${\cal F}_{\alpha_k}$ to all of $H$.

\item {\bf Strictly quasiclassical:} In a strictly quasiclassical UDM object, if $\alpha \leftarrow \beta, \alpha \neq \beta$ implies that ${\cal I}_\alpha \cup [ {\cal F}_\alpha ]  \subset {\cal I}_\beta$. 

\item {\bf Quasi-classical:} A UDM object is quasi-classical if it is sequential, and if $\alpha \leftarrow \beta$, then ${\cal I}_\alpha \subset {\cal I}_\beta$. 

\item {\bf Causal: } See Definition~\ref{causal}. 

\item {\bf Solvable: } See Definition~\ref{solvable}. 

\item {\bf Without self-information:} A UDM object has no self-information if for all its decision elements $\alpha \in A$, it holds that ${\cal I}_\alpha \subset {\cal F}(A - \{ \alpha \})$. 

\end{enumerate} 

A detailed study of the properties ensuing from this classification can be found in \citep{witsenhausen:75}, for example, systems with $T_0$ topologies are precisely those that induce a partial ordering on computing elements, and also define a sequential system. We will discuss some of these properties based on our generalization of the intrinsic model using category theory below.

\section{Functors, Natural Transformations, and the Yoneda Lemma} 

We now introduce some additional terminology from category theory, including the important idea of {\em functors} that map from one category to another, preserving the underlying structure of morphisms, {\em natural transformations} that map from one functor to another, and and finally one of most important results in category theory, the {\em Yoneda lemma} and how it can be used to construct representations of functors and associated universal representations. Our goal in the subsequent section is to use this machinery to construct universal representations of intrinsic models. 

\begin{definition} 
A {\bf covariant functor} $F: {\cal C} \rightarrow {\cal D}$ from category ${\cal C}$ to category ${\cal D}$ is defined as the following: 
\begin{itemize} 
    \item An object ${\cal F}X$ of the category ${\cal D}$ for each object $X$ in category ${\cal C}$.
    \item A morphism ${\cal F}f: {\cal F}X \rightarrow {\cal F}Y$ in category ${\cal D}$ for every morphism $f: X \rightarrow Y$ in category ${\cal C}$. 
   \item The preservation of identity and composition: ${\cal F} \ id_X = id_{{\cal F}X}$ and $({\cal F} g) ({\cal F} g) = {\cal F}(f g)$ for any composable morphisms $f: X \rightarrow Y, g: Y \rightarrow Z$. 
\end{itemize}
\end{definition} 

\begin{definition} 
A {\bf contravariant functor} $F: {\cal C} \rightarrow {\cal D}$ from category ${\cal C}$ to category ${\cal D}$ is defined exactly like the covariant functor, except all the mappings are reversed. In the contravariant functor ${\cal F}: C^{\mbox{op}} \rightarrow D$, every morphism $f: X \rightarrow Y$ is assigned the reverse morphism ${\cal F} f: {\cal F} Y \in {\cal F} X$ in category ${\cal D}$. 
\end{definition} 

Our goal is to construct covariant and contravariant functorial representations of intrinsic models. To this end, we introduce the following functors that will prove of value below: 

\begin{itemize} 
\item For every object $X$ in a category ${\cal C}$, there exists a covariant functor ${\cal C}(X, -): {\cal C} \rightarrow {\bf Set}$ that assigns to each object $Z$ in ${\cal C}$ the set of morphisms ${\cal C}(X,Z)$, and to each morphism $f: Y \rightarrow Z$, the pushforward mapping $f_*: {\cal C}(X,Y) \rightarrow {\cal C}(X, Z)$. 

\item For every object $X$ in a category ${\cal C}$, there exists a contravariant functor ${\cal C}(-, X): {\cal C}^{\mbox{op}} \rightarrow {\bf Set}$ that assigns to each object $Z$ in ${\cal C}$ the set of morphisms ${\cal C}(X,Z)$, and to each morphism $f: Y \rightarrow Z$, the pullback mapping $f^*: {\cal C}(Z, X) \rightarrow {\cal C}(Y, X)$. 
\end{itemize} 

From the above examples, it is now relatively straightforward to see how to define covariant and contravariant functors from the category of intrinsic models to the category of sets, but we need to develop a bit more machinery to understand the significance of these functorial representations. 

\begin{definition} 
Let ${\cal F}: {\cal C} \rightarrow {\cal D}$ be a functor from category ${\cal C}$ to category ${\cal D}$. If for all objects $X$ and $Y$ in ${\cal C}$, the map ${\cal C}(X, Y) \rightarrow {\cal D}({\cal F}X, {\cal F} Y)$, denoted as $f \mapsto {\cal F}f$ is
\begin{itemize}
    \item injective, then the functor ${\cal F}$ is defined to be {\bf faithful}. 
    \item surjective, then the functor ${\cal F}$ is defined to be {\bf full}.  
    \item bijective, then the functor ${\cal F}$ is defined to be {\bf fully faithful}. 
\end{itemize}

\end{definition} 

Our goal is to construct fully faithful functorial embeddings of intrinsic models, which gives us an embedding of intrinsic models into the category of sets. 

\subsection{Natural Transformations and the Yoneda Lemma} 

\begin{definition}
Given two functors ${\cal F}, {\cal G}: {\cal C} \rightarrow {\cal D}$ that map from category ${\cal C}$ to category ${\cal D}$, a {\em natural transformation} $\eta: {\cal F} \rightarrow {\cal G}$ consists of a morphism $\eta_X: {\cal F}X \rightarrow {\cal G} X$ for each object $X$ in ${\cal C}$. Moreover, these morphisms should satisfy the following property, that is the diagram below should commute: 
\begin{center}
    \begin{tikzcd}
  {\cal F}X \arrow[r, "{\cal F} f"] \arrow[d, "\eta_X" red]
    & {\cal F}Y \arrow[d, "\eta_Y" red] \\
  {\cal G}X  \arrow[r,  "{\cal G} f" ]
& {\cal G}Y
\end{tikzcd}
\end{center}
\end{definition}

\begin{definition}
For any two functors ${\cal F}, {\cal G}: {\cal C} \rightarrow {\cal D}$, let $\mbox{Nat}({\cal F}, {\cal G})$ denote the natural transformations from ${\cal F}$ to ${\cal G}$. If $\eta_X: {\cal F}X \rightarrow {\cal G} X$ is an isomorphism for each $X$ in category ${\cal C}$, then the natural transformation $\eta$ is called a {\bf natural isomorphism} and ${\cal F}$ and ${\cal G}$ are naturally isomorphic, denoted as ${\cal F} \cong {\cal G}$. 
\end{definition}

The machinery of natural transformations between functors enables making concrete the central philosophy underlying category theory, which is construct representations of objects in terms of their interactions with other objects. Unlike set theory, where an object like a set is defined by listing its elements, in category theory objects have no explicit internal structure, but rather are defined through the morphisms that they define with respect to other objects. The celebrated Yoneda lemma makes this philosophical statement more precise. 

\begin{theorem}
{\bf Yoneda Lemma:} For every object $X$ in category ${\cal C}$, and every contravariant functor ${\cal F}: {\cal C}^{\mbox{Op}} \rightarrow {\bf Set}$, the set of natural transformations from ${\cal C}(-, X)$ to ${\cal F}$ is isomorphic to ${\cal F} X$. 
\end{theorem}

That is, the natural transformations from ${\cal C}(-, X)$ to ${\cal F}$ serve to fully characterize the object ${\cal F} X$ up to isomorphism. In the special circumstance when the set-valued functor ${\cal F} = {\cal C}(-, Y)$, the Yoneda lemma asserts that $\mbox{Nat}({\cal C}(-, X), {\cal C}(-, Y) \cong {\cal C}(X, Y)$. In other words, a pair of objects are isomorphic $X \cong Y$ if and only if the corresponding contravariant functors are isomorphic, namely ${\cal C}(-, X) \cong {\cal C}(-, Y)$. 

\subsection{Presheaf Representations} 

A very important class of representations that follow from the Yoneda lemma are {\em presheafs} ${\cal C}(-, X)$. Given any two categories ${\cal C}, {\cal D}$, we can always define the new category ${\cal D}^{\cal C}$, whose objects are functors ${\cal C} \rightarrow {\cal C}$, and whose morphisms are natural transformations. If we take ${\cal D} = {\bf Set}$, and consider the contravariant version ${{\bf Set}^{\cal C}}^{\mbox{Op}}$, we obtain a category whose objects are presheafs. Presheafs have some very nice properties, which makes them a {\em topos} \citep{goldblatt}. 

Given a category of intrinsic models ${\cal C}_I$, or in particular a category of MDPs ${\cal C}_{\mbox{MDP}}$ with bisimulation homomorphism or a category of PSRs ${\cal C}_{PSR}$ with the defined PSR homomorphism, we can clearly apply the Yoneda lemma to construct presheaf representations of these decision making objects. A detailed study of each individual case is outside the scope of this introductory paper, and is a topic for future research. 

\section{Homotopical Representations of UDMs} 
\label{intr-homotopy}

We have seen that information fields induce a finite topological space over a UDM, enabling the decomposition of the UDM object into subsystems. In this section, we explore the topological ramifications of this idea further. Homotopy is a fundamental idea in algebraic topology, and we build on the use of homotopical constructions over finite topological spaces \citep{barmak}. A fundamental idea throughout mathematics is that of gleaning insight into the structure of one space by probing it with objects from another space. Thus, a fundamental way to understand the category of groups is to map it to the category of group representations. Similarly, in a topological space ${\bf Top}$, two objects $X$ and $Y$ are considered isomorphic if the corresponding sets ${\bf Top(Z, X)}$ and ${\bf Top(Z, Y)}$ are isomorphic. The Yoneda lemma described above allows us to construct functors from any category ${\cal C}$ to the category ${\bf Set}$. Our goal is to be able to compute invariant representations of UDM objects, such as their homotopies, and understand how to compute the {\em fundamental group} associated with a UDM object. We begin by reviewing some basic material on connectivity in finite topological spaces, and then show how various UDM objects can be compared in terms of their subsystem topologies.  

\subsection{Connectivity in UDMs} 

Since UDMs define a finite topological space, we can build on the core ideas of connectivity in such spaces \citep{barmak}. Every concept in a topological space must be defined in terms of the open (or closed) set topology, and that includes (path) connectivity. The crucial idea here is that connectivity is defined in terms of a continuous mapping from the unit interval $I = (0,1)$ to a topological space $X$. 

\begin{lemma}
A function $f: X \rightarrow Y$ between two finite spaces is continuous if and only if it is order-preserving, meaning if $x \leq x'$ for $x, x' \in X$, this implies $f(x) \leq f(x')$. 
\end{lemma}

\begin{definition}
We call two points $x, y \in X$ {\bf comparable} if there is a sequence of elements $x_0, \ldots, x_n$, where $x_0 = x, x_n = y$ and for each pair $x_i, x_{i+1}$ either $x_i \leq x_{i+1}$ or $x_i \geq x_{i+1}$. A {\bf fence} in $X$ is a sequence $x_0, x_1, \ldots, x_n$ of elements such that any two consecutive elements are comparable. $X$ is {\bf order connected} if for any two elements $x, y \in X$, there exists a fence starting in $x$ and ending in $y$. 
\end{definition}
 
 \begin{lemma}
 Let $x, y$ be two comparable points in a finite space $X$. Then, there exists a path from $x$ to $y$ in $X$, that is, a continuous map $\alpha: (0, 1) \rightarrow X$ such that $\alpha(0) = x$ and $\alpha(1) = y$. 
 \end{lemma}
 
 \begin{lemma}
 Let $X$ be a finite space. The following are equivalent: (i) $X$ is a connected topological space. (ii)  $X$ is an order-connected topological space (iii) $X$ is a path-connected topological space. 
 \end{lemma}
 
 A crucial strength of the topological perspective is the ability to combine two UDM objects  $X$ and $Y$ into a new space, which can generate a rich panoply of new objects. Here are a few of the myriad ways in which UDM objects can be combined \cite{munkres:algtop}. Table~\ref{example-enum} illustrates some of these ways of combining spaces for a small UDM object whose space of elements $X$ is comprised of just three agents. 

\begin{itemize} 
\label{prod-top}
\item Subspaces: The {\bf subspace} topology on $A \subset X$ is defined by the set of all intersections $A \cap U$ for open sets $U$ over $X$. 
\item Quotient topology: The {\bf quotient topology} on $U$ defined by a surjective mapping $q: X \rightarrow Y$ is the set of subsets $U$ such that $q^{-1}(U)$ is open on $X$. 
\item Union: The {\bf topology of the union} of two spaces $X$ and $Y$ is given by their disjoint union $X \bigsqcup Y$, which has as its open sets the unions of the open sets of $X$ and that of $Y$. 
\item Product of two spaces: The {\bf product topology} on the cartesian product $X \times Y$ is the topology with basis the ``rectangles" $U \times V$ of an open set $U$ in $X$ with an open set $V$ in $Y$.
\item Wedge sum of two spaces: The {\em wedge sum} is the ``one point" union of two ``pointed" spaces $(X, x_o)$ with $(Y, y_o)$, defined by $X \bigvee Y / x_0 \sim y_0$, the quotient space of the disjoint union of $X$ and $Y$, where $x_0$ and $y_0$ are identified. 
\item Smash product: The {\em smash product} topology is defined as the quotient topology $X \bigwedge Y = X \times Y / X \bigvee Y$. 
\item Non-Hausdorff cone: The {\bf non-Hausdorff cone} of topological space $X$ with $Y = \{ * \}$ yields the new space $\mathbb{C}(X)$, whose open sets are now ${\cal O}_{\mathbb{C}(X)} = {\cal O}_X \cup \{X \cup \{ * \} \}$. 
\item Non-Hausdorff suspension: The {\bf  non-Hausdorff suspension} of topological space $X$ with $Y = \{+, - \}$ yields the new space $\mathbb{S}(X)$, whose open sets are now ${\cal O}_{\mathbb{C}(X)} = {\cal O}_X \cup \{X \cup \ \{+, - \} \}$. 
\end{itemize} 
 
 \begin{table}
 \caption{Examples of UDM objects $A =\{a, b, c \}$ with different subsystem topologies. A proper open set is any set other than $\emptyset$ or $A$ (which are in any topology). $P_n$ is a topology on a set of size $n$ with only one proper open set. $D_n$ is the discrete topology over $n$ elements. $P_{m,n}$ are topologies where the proper open sets are all non-empty subsets of a subset of size $m$. The $\cong$ equivalence relation  is homotopy equivalence. See text for explanation.}
 \centering
 \begin{small}
  \begin{tabular}{|c|c|c|c|c|} \hline 
Proper Open Sets & Name & $T_0$? & Connected? & Equivalent graphical model \\ \hline 
All & $D_3$ & yes & no & HEDG (hyper-edge over (a,b,c))  \\ \hline 
$b, c$ & & yes & yes & DAG $b \rightarrow a$, $c \rightarrow a$ (collider over a) \\ \hline
$a, b, (a,b)$  & $P_{2,3} \cong \mathbb{C}D_2$  & yes &  yes & Chain graph: $a \rightarrow c, b \rightarrow c, a - b$ \\ \hline 
$a, b, (a,b), (b,c)$ & $D_1 \bigsqcup P_2$  & yes &  no & DAG with node $a$ disconnected, $b \rightarrow c$\\ \hline 

$a$  & $P_3$  & no &  yes &  Chain graph: $a \rightarrow b$, $a \rightarrow c$, and $b - c$\\ \hline
\end{tabular}
\end{small}
\label{example-enum}
\end{table}

\subsection{UDM Homotopies over Finite Topological Spaces} 

\begin{definition}
Let $f, g: X \rightarrow Y$ be two continuous maps between finite space topologies $X$ and $Y$. We say $f$ is {\em homotopic} to $g$, denoted as $f \cong g$ if there exists a continuous map $h: X \times [0,1] \rightarrow Y$ such that $h(x, 0) = f(x)$ and $h(x, 1) = g(x)$. In other words, there is a smooth ``deformation" between $f$ and $g$, so we can visualize $f$ being slowly warped into $g$. Note that $\cong$ is an equivalence relation, since $f \cong f$ (reflexivity), and if $f \cong g$, then $g \cong f$ (symmetry), and finally $f \cong g, g \cong h \ \ \implies f \cong h$ (transitivity). 
\end{definition}
 
 \begin{definition}
 A map $f: X \rightarrow Y$ is a {\em homotopy equivalence} if there exists another map $g: Y \rightarrow X$ such that $g \circ f \cong id_X$ and $f \circ g \cong id_Y$, where $id_X$ and $id_Y$ are the identity mappings on $X$ and $Y$, respectively. 
 \end{definition}

\begin{definition}
 A topological space $X$ is {\bf contractible} if the identity map $id_X: X \rightarrow X$ is homotopically equivalent to the constant map $f(x) = c$ for some $c \in X$. 
 \end{definition}
 
For example, any convex subset $A \subset \mathbb{R}^n$ is contractible. Let $f(x) = c, c \in A$ be the constant map. Define the homotopy $H: A \times I \rightarrow X$ as equal to $H(x,t) = t c + (1 - t) x$. Note that at $t=0$, we have $H(x,0) = x$, and that at $t=1$, we have $H(x,1) = c$, and since $A$ is a convex subset, the convex combination $t c + (1 - t) x \in A$ for any $t \in [0,1]$. 
 
 \begin{theorem}
 If $X$ is a finite topological space containing a point $y$ such that the only open (or closed) subset of $X$ containing $y$ is $X$ itself, then $X$ is contractible. In particular, the non-Hausdorff cone $\mathbb{C}(X)$ is contractible for any $X$. 
 \end{theorem}

{\bf Proof:} Let $Y = \{ * \}$ denote the space with a single element, $*$. Define the retraction mapping $r: X \rightarrow *$ by $r(x) = *$ for all $x \in X$, and define the {\em inclusion} mapping $i: Y \rightarrow X$ by $i(*) = y$. Clearly, $r \circ i  = id_{Y}.$ Define the homotopy $h: X \rightarrow I \rightarrow X$ by $h(x,t) = x$ if $t < 1$, and $h(x, 1) = y$. Then, $h$ is continuous, because for any open set $U$ in $X$, if $y \in U$, then clearly $U = X$ (as $X$ is the only open set containing $y$), and hence $h^{-1}(U) = X \times I$, which is open. If on the other hand, $y \notin U$, then $h^{-1}(U) = U \times [0, 1)$. It follows that $h$ is a homotopy $h \cong id_X = i \circ r$. \qed 

The following lemma is of crucial importance, showing how elements of a topological space that can be removed, reducing model size. 

\begin{definition}
A point $x$ in a finite  topological space $X$ is {\bf maximal} if there is no $y > x$, and {\bf minimal} if there is no $y < x$. 
\end{definition}
 
 \begin{lemma}
 If $X$ is an finite  space, then $U_x$ is contractible. In particular, if $X$ has a unique maximal point or unique minimal point, then $X$ is contractible. 
 \end{lemma}

\begin{definition}
Let $f, g: X \rightarrow Y$ be two continuous maps between finite space topologies $X$ and $Y$. We say $f$ is {\em homotopic} to $g$, denoted as $f \cong g$ if there exists a continuous map $h: X \times [0,1] \rightarrow Y$ such that $h(x, 0) = f(x)$ and $h(x, 1) = g(x)$. In other words, there is a smooth ``deformation" between $f$ and $g$, so we can visualize $f$ being slowly warped into $g$. Note that $\cong$ is an equivalence relation, since $f \cong f$ (reflexivity), and if $f \cong g$, then $g \cong f$ (symmetry), and finally $f \cong g, g \cong h \ \ \implies f \cong h$ (transitivity). 
\end{definition}
 
 \begin{definition}
 A map $f: X \rightarrow Y$ is a {\em homotopy equivalence} if there exists another map $g: Y \rightarrow X$ such that $g \circ f \cong id_X$ and $f \circ g \cong id_Y$, where $id_X$ and $id_Y$ are the identity mappings on $X$ and $Y$, respectively. 
 \end{definition}

\subsection{Efficient Enumeration of Homeomorphically Distinct UDMs} 
\label{homotopy}

Next, we turn to the fundamental problem of how to construct homemorphically distinct UDMs. In the definitions below, we focus purely on the topological structure of a UDM, namely the subsystem topology as defined in Definition~\ref{subsystem-defn}.  
\begin{definition}
For every object ${\cal M}$ in a UDM with $T_0$ subsystem topology that defines a partial ordering $\leq$, define its associated {\bf Hasse diagram} $H_{\cal M}$ as a directed graph which captures all the relevant order information of ${\cal M}$. More precisely, the  vertices of $H_{\cal M}$ are the elements of ${\cal M}$, and the edges of $H_{\cal M}$ are such that there is a directed edge from $x$ to $y$ whenever $y \leq x$,  but there is no other vertex $z$ such that
$y \leq  z \leq x$.
\end{definition}

General pre-ordered UDM objects  can be reduced to partially ordered  UDM objects with $T_0$ topologies up to homomeomorphic equivalence. 

\begin{theorem}\citep{stong}
Let $(X, \cal{T})$ be an arbitrary UDM object  with a subsystem topology defining an associated preordering $\leq$. Let $X_0$ represent the quotient topological space $X / \sim$, where $x \sim y$ if $x \leq y$ and $y \leq x$. Then $X_0$ is a homotopically equivalent intrinsic  model with $T_0$ separability, and the quotient map $q: X \rightarrow X_0$ is a homotopy equivalence. Furthermore, $X_0$ induces a partial ordering on the elements $x \in X_0$. 
\end{theorem}

A key idea in the enumeration is to assume that each element in the Hasse diagram of the poset does not have an in-degree or out-degree of $1$. 
\begin{definition} \citep{stong}
An element $x \in X$ in a UDM object with $T_0$  subsystem topology $X$ is a {\bf down beat} point if $x$ covers one and only one element of of $X$. Alternatively, the set $\hat{U}_x = U_x \setminus \{x\}$ has a (unique) maximum. Similarly, $x \in X$ is an {\bf up beat} point if $x$ is covered by a unique element, or equivalently if $\hat{F}_x = F_x \setminus \{ x \}$ has a (unique) minimum. A {\bf beat} point is either a down beat or up beat point. 
\end{definition}
\begin{definition}
A subspace $A \subset X$ is called a strong deformation retract of $X$ if there is a homotopy $F: X \times [0,1] \rightarrow A$ such that $F(x,0) = x, F(x, 1) \in A, F(a,t) = a$ for all $x \in X, t \in [0,1], a \in A$.
\end{definition}
\begin{theorem}\citep{stong}
Let $X$ be a UDM object with $T_0$ subsystem topological model, and let $x \in X$ be a (down, up) beat point. Then the reduced object $X \setminus \{ x \}$ is a {\bf strong deformation retract} of $X$. An element  $x$ in a UDM object   ${\cal M}$ is an upbeat point if and only if it has in-degree one in the associated Hasse diagram $H_{\cal M}$, i.e., it has only one incoming
edge). Similarly, $x$ is downbeat if and only if it has out-degree one (it has only one
outgoing edge).
\end{theorem} 

\begin{definition}
A UDM object with a  $T_0$ subsystem topological space is a {\bf minimal} if it has no beat points. A {\bf core} of a UDM object  $X$ is a strong deformation retract, which is a minimal finite space. The {\bf minimal graph} of a minimal UDM object is its equivalent Hasse diagram. 
\end{definition}

\begin{theorem}\citep{stong}
{\bf Classification Theorem:} A homotopy equivalence between minimal UDM objects is a homeomorphism. In particular, the core of a UDM object is unique up to homeomorphism and two UDM objects  are homotopy equivalent if and only if they have homeomorphic cores. 
\end{theorem}

\begin{figure}
 \caption{Left: Constructing minimal UDM objects  by removing beat points \citep{barmak,stong}. Right: Efficiently enumerating minimal UDM objects (based on the enumeration method in \citep{fix-patrias}).}
  \begin{minipage}[t]{.2\linewidth}
\vspace{0pt}
\centering
\includegraphics[scale=0.25]{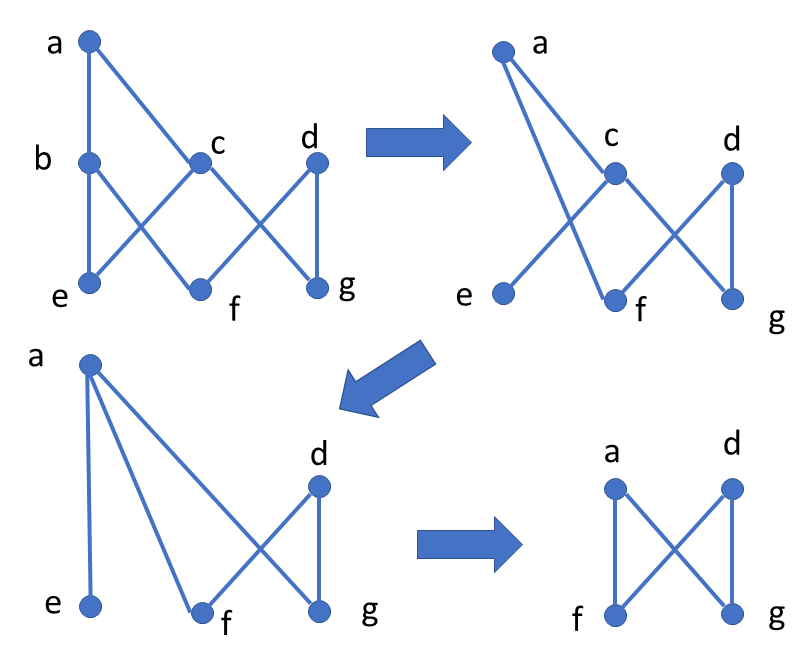} 
\end{minipage} \hfill 
 \begin{minipage}[t]{.5\linewidth}
\vspace{0pt}
\centering
\includegraphics[scale=0.13]{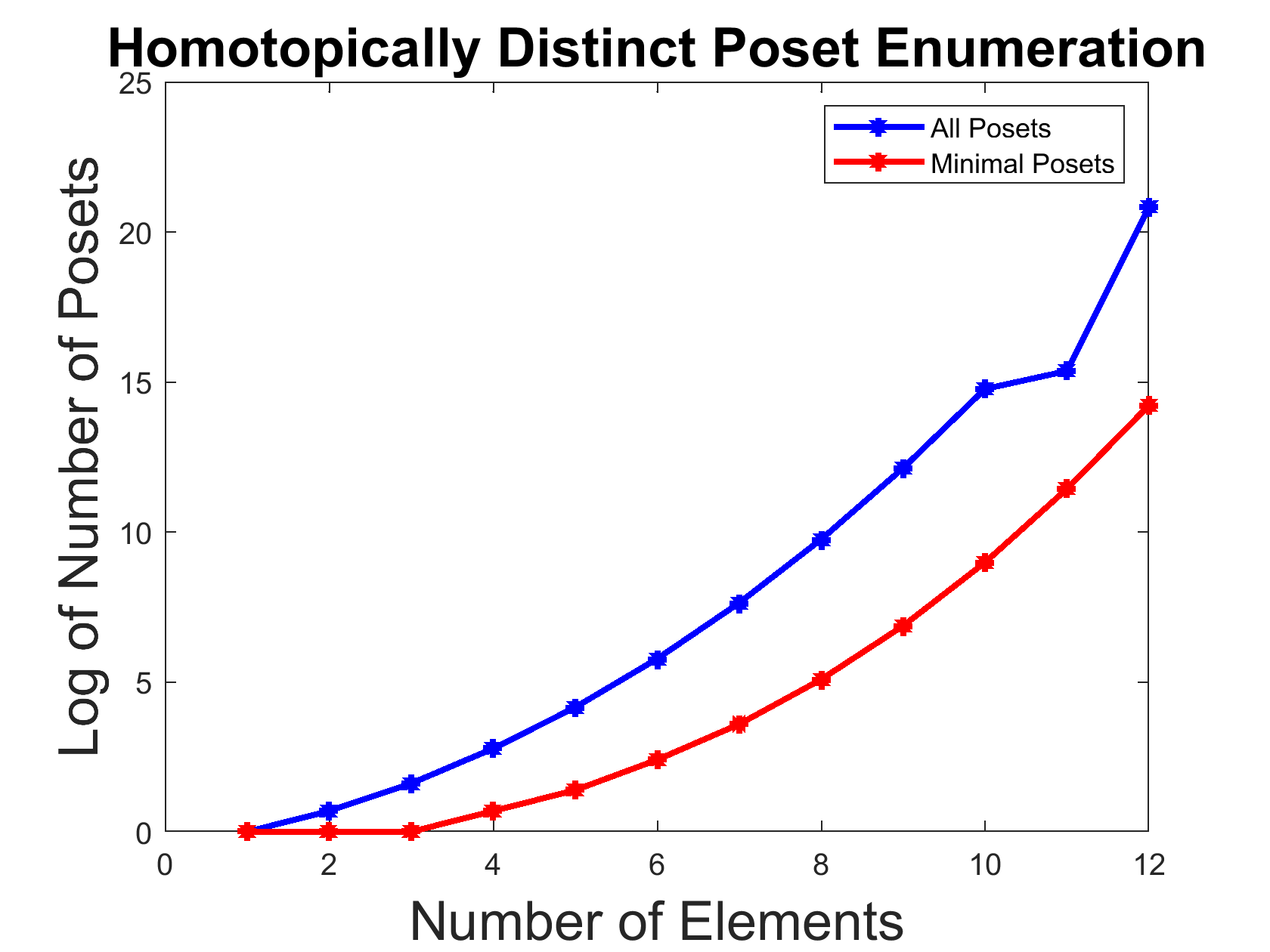} 
\end{minipage} 
 \label{enum-t0}
\end{figure}

Figure~\ref{enum-t0} illustrates the process of removing beat points to construct the minimal UDM object. b is an up beat point of $X$, c is an upbeat point of $X \setminus \{b \}$, and e is an up beat point of $X \setminus \{b, c\}$. Similarly, points c and e are removed, resulting in the minimal object. The figure also shows that homeomorphic equivalences greatly reduces the search space of possible object structures. Note the plot is on log scale.  For example, for $12$ variables, the number of minimal objects is $< 0.1$\% of the number of possible objects, a savings of three orders of magnitude. 

\begin{algorithm}[t]
\caption{Find Topologically Minimal UDM object  with $T_0$ Subsystem Topology.}
\SetAlgoLined
\KwIn{	General UDM object ${\cal M}$ with a  reflexive transitive pre-ordered structure induced by the subsystem topology.\\
}
\KwOut{Minimal UDM object with $T_0$ subsystem  topology homotopically equivalent to original pre-ordered object. \\
		The algorithm uses homotopy theory to find the {\em core} $T_0$ description of a pre-ordered UDM object.}
\Begin{
    Define the topology $(X, \mathcal{U})$ where $X = V$ and the open sets in $\mathcal{U}$ are constructed from the induced pre-order $\leq$ from ${\cal M}$. 
	Define the minimal object $(X_0, \mathcal{U'})$, and set $X_0 = X$. \\
	\Repeat{convergence}{
	\For{$x,  y \in X_0$ s.t. $x \leq y, y \leq x$}{
	 Remove $x, y$ from $X_0$, and replace them with a new variable $z = x \sim y$. \\
	 Set $X_0 \leftarrow X_0 \setminus \{ x, y \} \cup \{ z \}$. $z$ represents the equivalence class that includes $x$ and $y$. \\
	 }
	 \For{$x \in X_0$}{
	 {\bf Remove down beat points:} If $\hat{U}_{x} = U_x \setminus \{ x \}$ has a maximum, then $X_0 \leftarrow X_0 \setminus \{ x \}$.\\ 
	 {\bf Remove up beat points:} If $\hat{F}_{x} = F \setminus \{ x \}$ has a minimum, then $X_0 \leftarrow X_0 \setminus \{ x \}$. 
	 }
	 }
	
	Define the open sets $U_x \in \mathcal{U'}$ as $U_x = \{y \ | \ y \leq x \}$ for $x \in X_0$. 
}
\end{algorithm}

Algorithm 1 determines a reduced UDM object based on discovering a quotient $T_0$ subsystem topology that is homotopically equivalent to original  non-reduced object with a non-$T_0$ topology. Second, the algorithm further reduces the object to its core by removing {\em beat points} \citep{barmak,stong}. 
  
\section{Equilibration in UDMs} 

At the outset, we cautioned that the bulk of this paper is devoted to a study of information structures that underly decision making broadly, which precludes introducing particular solution methodologies, such as dynamic programming \citep{DBLP:books/lib/Bertsekas05}, which is specific to sequential information structures. In this section, however, we discuss a broad solution methodology called {\em equilibration}, which applies to multiplayer network games, reinforcement learning, and to causal inference, based on a generalization of optimization called variational inequalities (VIs) \citep{nagurney:vibook}. We use a running example from earlier in the paper of a producer consumer multiplayer game, as shown in Figure~\ref{soi}, and analyze its solution in depth in this section. VIs can be seen as a generalization of optimization, and frequently used to solve complex network games. This section is a condensed version of a recent paper on causal VIs \citep{causal-vi}, which the reader is encouraged to read for further details. 

\subsection{Causal Variational Inequalities}
\label{cvi} 

Our variational formulation of causal inference  is a synthesis of classical variational inequalities \citep{facchinei-pang:vi} and causal models \citep{rubin-book,pearl-book}.  More precisely, a causal variational inequality model $\cal{M} =$ CVI($F,K$), where $F$ is a collection of modular vector-valued functions defined as $F_i$, where $F_i: K_i \subset \mathbb{R}^{n_i} \rightarrow \mathbb{R}^{n_i}$, with each $K_i$ being a convex domain such that $\prod_i K_i = K$. We assume that the domains of each $F_i$ range over a collection $V$ of endogenous variables, and a set $U$ of exogenous variables, where only the endogenous variables are subject to causal manipulation. We model each intervention as a submodel $F_w$, and each component of $F_w$ reflects the effect of some manipulation of a subset $V_w \subseteq V$ of endogenous variables. 

\begin{definition}
\label{cvidef}
The category ${\cal C}_{\mbox{CVI}}$ of causal VIs is defined as one where each object is defined  as a 
finite-dimensional causal  variational inequality problem ${\cal M}$ = CVI($F, K)$, where the vector-valued mapping $F$ depend on both deterministic and stochastic elements, namely $F(x) = E[F(x, \eta)]$. where $\eta$ is a random variable defined over the probability space $(\Omega, {\cal F}, P)$, $E[.]$ denotes expectation with respect to the probability distribution $P$ over the random variable $\eta$, and $F: K \rightarrow \mathbb{R}^n$ is a given continuous function, $K$ is a given closed convex set, and $\langle .,.\rangle$ is the standard inner product in $\mathbb{R}^n$. A causal intervention is modeled as a submodel ${\cal M}_w$ = CVI($F_w, K)$, where $F_w(x) = E_w[F(x, \eta | \hat{w})]$, where $\hat{w}$ denotes the intervention of setting of variable $w$ to a specific non-random value, and where $E_w[.]$ now denotes expectation with respect to the intervention probability distribution $P_w$. Solving a causal VI is defined as finding a vector $x^* = (x^*_1, \ldots, x^*_n) \in K \subset \mathbb{R}^n$ such that
\begin{equation*}
\langle F_w(x^*), (y - x^*) \rangle \geq 0, \ \forall y \in K
\end{equation*}
\end{definition}
\begin{figure}[h]
\begin{center}
\begin{minipage}[t]{0.45\textwidth}
\includegraphics[width=\textwidth,height=1.25in]{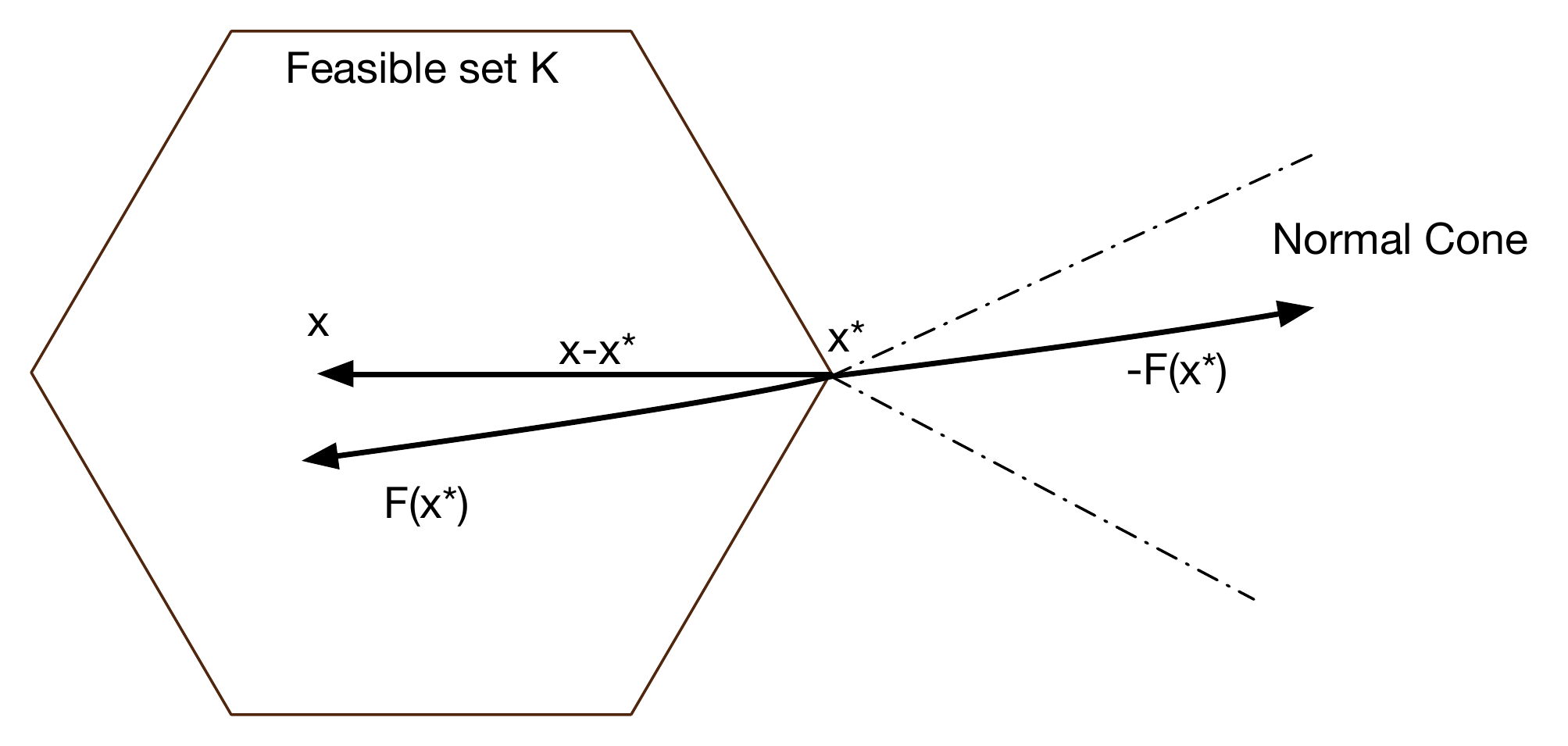}
\end{minipage}
\end{center}
\caption{This figure provides a geometric interpretation of a causal variational inequality $CVI(F_w,K)$. The mapping $F_w$ defines a vector field over the feasible set $K$ and a probability space, where $E_w(F(x, \eta) | \hat{w})$ is the conditional mean vector field (denoted in the figure by $F$), computed over the intervention distribution $P_w$. At the solution point $x^*$, the vector field $F(x^*)$ is directed inwards at the boundary, and  $-F(x^*)$ is an element of the normal cone $C(x^*)$ of $K$ at $x^*$ where the normal cone  $C(x^*)$ at the vector $x^*$ of a convex set $K$ is defined as $C(x^*) = \{y \in \mathbb{R}^n | \langle y, x - x^* \rangle \leq 0, \forall x \in K \}$.}
\label{vi-geom}
\end{figure}

\subsection{Properties of Mappings} 
The solution to a (causal) VI depends on the properties satisfied by the mapping $F$ and the feasible space $K$. If $K$ is compact and $F$ is continuous, it is straightforward to prove using Brower's fixed point theorem that there is always at least one solution to any VI (see Theorem~\ref{projthm}).  However, to obtain a unique solution, a stricter condition is necessary. 
\begin{definition}
$F(x)$ is {\em monotone} if $\langle F(x) - F(y), x - y \rangle \ge 0$, $\forall x, y \in K$.
\end{definition}

\begin{definition}
$F(x)$ is {\em strongly monotone} if $\langle F(x) - F(y), x  - y \rangle \geq \mu \| x - y \|^2_2, \mu > 0, \forall x,y \in K$.
\end{definition}
\begin{definition}
$F(x)$ is {\em Lipschitz} if $\| F(x) - F(y) \|_2 \leq L \|x - y \|_2, \forall x,y \in K$.
\end{definition}

Crucially, VI problems can only be converted into equivalent optimization problems when a very restrictive condition is met on the Jacobian of the mapping $F$, namely that it be symmetric. Most often, real-world applications of VIs, such as the example in Section~\ref{soi}, do not induce symmetric Jacobians. 
\begin{theorem}
\label{equivalence}
Assume $F(x)$ is continuously differentiable on $K$ and that the Jacobian matrix $\nabla F(x)$ of partial derivatives of $F_i(x)$ with respect to (w.r.t) each $x_j$ is symmetric and positive semidefinite. Then there exists a real-valued convex function $f: K \rightarrow \mathbb{R}$ satisfying $\nabla f(x) = F(x)$ with $x^*$, the solution of  VI(F,K), also being the mathematical programming problem of minimizing $f(x)$ subject to $x \in K$.
\end{theorem}
The algorithmic development of methods for solving VIs begins with noticing their connection to fixed point problems.
\begin{theorem}
\label{projthm}
The vector $x^*$ is the solution of VI(F,K) if and only if, for any $\alpha > 0$, $x^*$ is also a fixed point of the map  $x^* = P_K(x^* - \alpha F(x^*))$,
where $P_K$ is the projector onto convex set $K$.
\end{theorem}
In terms of the geometric picture of a VI illustrated in Figure \ref{vi-geom}, this property means that the solution of a VI occurs at a vector $x^*$ where the vector field $F(x^*)$ induced by $F$ on $K$ is normal to the boundary of $K$ and directed inwards, so that the projection of $x^* - \alpha F(x^*)$ is the vector $x^*$ itself. This property forms the basis for the projection class of methods that solve for the fixed point.

\subsection{Causal Variational Inequalities and Games}
\label{theory_nash}
VIs are a mathematically elegant approach to modeling and solving equilibrium problems in game theory \citep{game-theory-book}.  A {\em Nash game} consists of $m$ players, where player $i$ chooses a strategy $x_i$ belonging to a closed convex set $X_i \subset \mathbb{R}^n$.  After executing the joint action, each player is penalized (or rewarded) by the amount $f_i(x_1,\ldots,x_m)$, where $f_i: \mathbb{R}^n \rightarrow \mathbb{R}$ is a continuously differentiable function.  A set of strategies $x^* = (x_1^*,\ldots,x_m^*) \in \Pi_{i=1}^M X_i$ is said to be in equilibrium if no player can reduce the incurred penalty (or increase the incurred reward) by unilaterally deviating from the chosen strategy.  If each $f_i$ is convex on the set $X_i$, then the set of strategies $x^*$ is in equilibrium if and only if $\langle \nabla_i f_i (x_i^*), (x_i - x_i^*) \rangle \ge 0$.  In other words, $x^*$ needs to be a solution of the VI $\langle F(x^*), (x-x^*) \rangle \ge 0$, where $F(x) = ( \nabla f_1(x), \ldots, \nabla f_m(x))$.

Complementarity problems provide the foundation for a number of Nash equilibrium algorithms.  The class of complementarity problems can also be reduced to solving a VI. When the feasible set $K$ is a cone, meaning that if $x \in K$, then $\alpha x \in K, \alpha \geq 0$, then the VI becomes a CP.
\begin{definition}
Given a cone $K \subset \mathbb{R}^n$ and mapping $F: K \rightarrow \mathbb{R}^n$, the complementarity problem CP(F,K) is to find an $x \in K$ such that $F(x) \in K^*$, the dual cone to $K$, and $\langle F(x), x \rangle \geq 0$. \footnote{Given a cone $K$, the dual cone $K^*$ is defined as $K^* = \{ y \in \mathbb{R}^n | \langle y, x \rangle \geq 0, \forall x \in K \}$.}
\end{definition}

The nonlinear complementarity problem (NCP) is to find $x^* \in \mathbb{R}^n_+$ (the non-negative orthant) such that $F(x^*) \geq 0$ and $\langle F(x^*), x^* \rangle = 0$. The solution to an NCP and the corresponding $VI(F, \mathbb{R}^n_+)$ are the same, showing that NCPs reduce to VIs. In an NCP, whenever the mapping function $F$ is affine, that is $F(x) = Mx + b$, where $M$ is an $n \times n$ matrix, the corresponding NCP is called a linear complementarity problem (LCP) \citep{murty:lcpbook}.

\subsection{Causal Network Economics}

We now describe how to model causal inference in a network economics problem, which will be useful in illustrating the abstract definitions from the previous section. The model in Figure~\ref{soi} is drawn from \citep{nagurney:vibook,nagurney:soi}. which were  deterministic, and included no analysis of causal interventions. This network economics model comprises of three tiers of agents: producer agents, who want to sell their goods, transport agents who ship merchandise from producers, and demand market agents interested in purchasing the products or services. The model applies both to electronic goods, such as video streaming, as well as physical goods, such as face masks and other PPEs. Note that the design of such an economic network requires specifying the information fields for every producer, transporter and consumer. For the sake of brevity, we assume that the definition of these information fields are implicit in the equations defined below, but a fuller discussion of this topic will be studied in a subsequent paper. 

The model assumes $m$ service providers, $n$ network providers, and $o$ demand markets.  Each firm's utility function is defined in terms of the nonnegative service quantity (Q), quality (q), and price ($\pi$) delivered from service provider $i$ by network provider $j$ to consumer $k$.  Production costs, demand functions, delivery costs, and delivery opportunity costs are designated by $f$, $\rho$, $c$, and $oc$ respectively.  Service provider $i$ attempts to maximize its utility function $U_i^1(Q,q^*,\pi^*)$ by adjusting $Q_{ijk}$ (eqn. \ref{U1}).  Likewise, network provider $j$ attempts to maximize its utility function $U_j^2(Q^*,q,\pi)$ by adjusting $q_{ijk}$ and $\pi_{ijk}$ (eqn. \ref{U2}).

\begin{subequations}
\begin{align}
\label{U1}
U_i^1(Q,q^*,\pi^*) &= \sum_{j=1}^n \sum_{k=1}^o \hat{\rho}_{ijk}(Q,q^*)Q_{ijk} - \hat{f}_i(Q)\\
&- \sum_{j=1}^n \sum_{k=1}^o \pi^*_{ijk}Q_{ijk}, \hspace{0.2cm} Q_{ijk} \ge 0 \nonumber
\end{align}
\begin{align}
\label{U2}
U_j^2(Q^*,q,\pi) = &\sum_{i=1}^m \sum_{k=1}^o \pi_{ijk}Q^*_{ijk}\\
- &\sum_{i=1}^m \sum_{k=1}^o (c_{ijk}(Q^*,q) + oc_{ijk}(\pi_{ijk})), \nonumber \\
&q_{ijk}, \pi_{ijk} \ge 0 \nonumber
\end{align}
\end{subequations}

We assume the governing equilibrium is Cournot-Bertrand-Nash and the utility functions are all concave and fully differentiable.  This establishes the equivalence between the equilibrium state we are searching for and the variational inequality to be solved where the $F$ mapping is a vector consisting of the negative gradients of the utility functions for each firm.  Since $F$ is essentially a concatenation of gradients arising from multiple independent, conflicting objective functions, it does not correspond to the gradient of any single objective function.  

\begin{subequations}
\begin{align}
\label{SOI-vi}
&\langle F(X^*),X-X^* \rangle \ge 0, \forall X \in \mathcal{K},\\
\text{where } \hspace{0.2cm} &X = (Q,q,\pi) \in \mathbb{R}^{3mno+} \nonumber \\
\text{and } \hspace{0.35cm} &F(X) = (F^1_{ijk}(X), F^2_{ijk}(X), F^3_{ijk}(X)) \nonumber
\end{align}
\begin{align}
F^1_{ijk}(X) &= \frac{\partial f_i (Q)}{\partial Q_{ijk}} + \pi_{ijk} - \rho_{ijk} - \sum_{h=1}^n \sum_{l=1}^o \frac{\partial \rho_{ihl} (Q,q) }{\partial Q_{ijk}} \times Q_{ihl} \label{F1} \\
F^2_{ijk}(X) &= \sum_{h=1}^m \sum_{l=1}^o \frac{\partial c_{hjl} (Q,q)} {\partial q_{ijk}} \label{F2} \\
F^3_{ijk}(X) &= -Q_{ijk} + \frac{\partial oc_{ijk}(\pi_{ijk})}{\partial \pi_{ijk}} \label{F3}
\end{align}
\end{subequations}

The variational inequality in Equations~\ref{SOI-vi} represents the result of combining the utility functions of each firm into standard form.  $F^1_{ijk}$ is derived by taking the negative gradient of $U_i^1$ with respect to $Q_{ijk}$.  $F^2_{ijk}$ is derived by taking the negative gradient of $U_j^2$ with respect to $q_{ijk}$.  And $F^3_{ijk}$ is derived by taking the negative gradient of $U_j^2$ with respect to $\pi_{ijk}$.

\subsubsection{Numerical Example} 

We extend the simplified numerical example in \citep{nagurney:soi} by adding stochasticity to illustrate our causal variational formalism. Let us assume that there are two service providers, one transport agent, and two demand markets. Define the production cost functions:
\[ f_1(Q) = q^2_{111} + Q_{111}  + \eta_{f_1}, f_2(Q) = 2 Q^2_{111} + Q_{211} + \eta_{f_2} \] 
where $\eta_{f_1}, \eta_{f_2}$ are random variables indicating errors in the model. 
Similarly, define the demand price functions as:
\begin{eqnarray*}
 \rho_{111}(Q,q) = -Q_{111} - 0.5 Q_{211} + 0.5 q_{111} + 100 + \eta_{\rho_{111}} \\
 \rho_{211}(Q,q) = -Q_{211} - 0.5 Q_{111} + 0.5 q_{211} + 200 + \eta_{\rho_{211}}
 \end{eqnarray*}
Finally, define the transportation cost functions as:
\begin{eqnarray*}
c_{111}(Q,q) = 0.5(q_{111} - 20)^2 + \eta_{c_{111}} \\ 
c_{211}(Q,q) = 0.5(q_{211} - 10)^2 + \eta_{c_{211}} 
\end{eqnarray*}
and the opportunity cost functions as:
\[ oc_{111}(\pi_{111}) = \pi_{111}^2 + \eta_{oc_{111}}, oc_{211}(\pi_{211}) =\pi_{211}^2 + \eta_{oc_{211}} \] 
Using the above equations, we can easily compute the component mappings $F_i$ as follows: 
\begin{eqnarray*}
F^1_{111}(X) = 4 Q_{111} + 0.5 Q_{211} - 0.5 q_{111} - 99 \\
F^1_{211}(X) = 6 Q_{211} + \pi_{211} - 0.5 Q_{111} - 0.5 q_{211} -199\\
F^2_{111}(X) = q_{111} - 20, \ F^2_{211}(X) = q_{211} - 10 \\
F^3_{111}(X) = -Q_{111} + 2 \pi_{111}, \ F^3_{211}(X) = -Q_{211} + 2 \pi_{211}
\end{eqnarray*}
For simplicity, we have not indicated the noise terms above, but assume each component mapping $F_i$ has an extra noise term $\eta_i$. It is also clear that we can now give precise semantics to causal intervention in this system, following the principles laid out in \citep{pearl-book}. For example, if we set the network service cost $q_{111}$ of network provider 1 serving the content producer 1 to destination market 1 to 0, then the production cost function under the intervention distribution is given by 
\[ E_{q_{111} = 0}(f_1(Q) | \hat{q}_{111}) = Q_{111} + E_{q_{111}=0} (\eta_{f_1} | \hat{q}_{111})\] 
Finally, the Jacobian matrix associated with $F(X)$ is given by the partial derivatives of each $F_i$ mapping with respect to $(Q_{111}, Q_{211}, q_{111}, q_{211}, \pi_{111}, \pi_{211})$ is given as:
\[ - \nabla U(Q,q,\pi) =  \left( \begin{array}{cccccc} 4 & .5 & -.5 & 0 & 1 & 0 \\
0.5 & 6 & 0 & -.5 & 0 & 1 \\ 0 & 0 & 1 & 0 & 0 & 0 \\ 0 & 0 & 0 & 1 & 0 & 0 \\ -1 & 0 & 0 & 0 & 2 & 0 \\ 0 & -1 & 0 & 0 & 0 & 2\end{array} \right) \]
Note this Jacobian is non-symmetric, but positive definite, as it is diagonally dominant. Hence the induced vector field $F$ can be shown to be strongly monotone, and the induced VI has exactly one solution. 

\subsection{Causal VI Algorithms} 

 We now discuss algorithms for solving causal VI's. There are a wealth of existing methods for deterministic VI's \citep{facchinei-pang:vi,nagurney:vibook}), which can be adapted to solving causal VI's.   The simplest method for solving a causal VI is the well-known projection algorithm \citep{facchinei-pang:vi}: 
\[ x_{k+1} = \Pi_K [x_k - \alpha_k F_w(x_k)] \]
where $F_w$ is the vector field induced by some causal intervention, 
which can be viewed as a modification of the classical projection method for deterministic VI's. The algorithm follows the direction of the negative vector field at a point $x_k$, and if the iterate falls outside the feasible space $K$, it projects back into $K$. If $F_w$ is strongly monotone, and Lipschitz, and the learning rate $\alpha_k$ is suitably designed, then the projection algorithm is guaranteed to find the solution to a causal VI.

Understanding the convergence of the projection method will give us insight into how to analyze causal interventions in VI's. At the heart of convergence analysis of any VI method is bounding the iterates of the algorithm. In the below derivation, $x^*$ represents the final solution to a causal VI, and $x_{k+1}$, $x_k$ are successive iterates: 
\begin{small}
\begin{eqnarray*}
\| x_{k+1} - x^* \|^2 &=& \| P_K[x_k - \alpha_k F_w(x_k)] - P_K[x^* - \alpha_k F_w(x^*)] \|^2 \\
&\leq& \| (x_k - \alpha_k F_w(x_k)) - (x^* - \alpha_k F_w(x^*)) \|^2 \\
&=& \| (x_k - x^*) - \alpha_k (F_w(x_k) - F_w(x^*)) \|^2 \\
&=& \|x_k - x^* \|^2 - 2 \alpha_k \langle (F_w(x_k) - F_w(x^*)), x_k - x^* \rangle \\ &+& \alpha_k^2 \|F_w(x_k) - F_w(x^*)\|^2 \\
&\leq& (1 - 2 \mu \alpha_k + \alpha_k^2 L^2) \| x_k - x^*\|^2
\end{eqnarray*} 
\end{small}
Here, the first inequality follows from the nonexpansive property of projections, and the last inequality follows from strong monotonicity and Lipschitz property of the $F_w$ mapping. Bounding the term $\langle (F_w(x_k) - F_w(x^*)), x_k - x^* \rangle$ is central to the design of any VI method. As we show in the next section, in modeling causal interventions a similar term will arise, except under different mappings, representing the ``untreated" and "treated" cases. 

\citet{korpelevich} extended the projection algorithm with the well-known ``extragradient" method, which requires two projections, but is able to solve VI's for which the mapping $F$ is only monotone.  If projections are expensive, particularly in large network economy models, these algorithms may be less attractive than incremental stochastic projection methods, which we turn to next. 

\subsection{Incremental Projection Methods} 

We now describe an incremental two-step projection method for solving causal VI's, based on work by \citet{DBLP:journals/mp/WangB15}. Their algorithm adapted to causal VI's can be written as follows:
\begin{equation}
\label{2stepalg}
z_k = x_k - \alpha_k F_w(x_k, v_k), \ \ \ x_{k+1} = z_k - \beta_k (z_k - P_{w_k} z_k)
\end{equation}
where $\{v_k\}$ and $\{w_k\}$ are sequences of random variables, generated by sampling the causal VI model, and $\{\alpha_k\}$ and $\{\beta_k\}$ are sequences of positive scalar step sizes. Note that an interesting feature of this algorithm is that the sequence of iterates $x_k$ is not guaranteed to remain within the feasible space $K$ at each iterate. Indeed, $P_{w_k}$ represents the projection onto a randomly sampled constraint $w_k$. 

The analysis of convergence of this algorithm is somewhat intricate, and we refer the reader to \citep{causal-vi} for more details. It can be shown that two-step stochastic algorithm given in Equation~\ref{2stepalg} converges to the solution of a causal VI, namely:
\begin{theorem}
Given the category of ${\cal C}_{\mbox{CVI}}$ causal VIs, the solution associated with any decision object representing a finite-dimensional causal  variational inequality problem  ${\cal M}$ = CVI($F, K)$, and a causal intervention, defined by the sub-object ${\cal M}_w$ = CVI($F_w, K)$, where $F_w(x) = E_w[F(x, \eta | \hat{w})]$, where $\hat{w}$ denotes the intervention of setting of variable $w$ to a specific non-random value, and where $E_w[.]$ now denotes expectation with respect to the intervention probability distribution $P_w$, the two-step algorithm given by Equation~\ref{2stepalg} produces a sequence of iterates $x_k$ that converges almost surely to $x^*$, where 
\begin{equation*}
\langle F_w(x^*), (y - x^*) \rangle \geq 0, \ \forall y \in K
\end{equation*}
\end{theorem}
{\bf Proof:} The proof of this theorem is given in \citep{causal-vi}, and largely follows the derivation given in \citep{DBLP:journals/mp/WangB15}, where the only difference is that in a causal VI problem, we are conditioning the stochastic VI on the intervention distribution $P_w$. $\qed$

\subsection{Decomposition of Causal VI UDMs} 

Finally, we discuss the issue of how to exploit structure in solving complex UDMs, specifically for the type of network economics problems described above. We also discuss ``sensitivity analysis" of UDMs, in particular,  how to measure the effect of some intervention, by comparing potential outcomes across the ``treated" units with the ``untreated" units \citep{rubin-book}. We characterize treatment effects in causal variational inequalities under interventions, building on the existing results on sensitivity analysis of classical variational inequalities \citep{nagurney:vibook}. 

We can apply the more general machinery of finite space topologies introduced above, but for simplicity, we focus on the case when a UDM object is defined by a cartesian product operation over the set of feasible solutions, and the monotone operator $F$ decomposes additively over these individual subsets. 
\begin{definition}
\label{partioned-cvi}
A partitioned CVI is defined as the causal  variational inequality problem of finding a vector $x^* = (x^*_1, \ldots, x^*_n) \in K \subset \mathbb{R}^n$ such that
\begin{equation*}
\langle E_w[F(x, \eta | \hat{w})], (x - y) \rangle \geq 0, \ \forall y \in K
\end{equation*}
where the function $F$ is partitionable function of order $m$, meaning that
\begin{equation*}
\langle E_w[F(x, \eta | \hat{w})], (x - y) \rangle = \sum_{i=1}^m \langle E_w[F_i(x, \eta | \hat{w})], (x_i - y_i) \rangle 
\end{equation*}
where each $F_i: K_i \subset \mathbb{R}^{n_i} \rightarrow \mathbb{R}^{n_i}$, with each $K_i$ being a convex domain such that $\prod_i K_i = K$.
\end{definition}
We can further simplify the solution of causal VIs by doing sensitivity analysis of a UDM subject to an intervention. The use of interventions to probe a structure is very common in many engineering domains, as well as more abstractly in many areas of math. 

\begin{theorem}
If $Y$ is probabilistically causally irrelevant to $X$, given $Z$, then  CVI$(F_{\hat{y},\hat{z}}(x), K)$ has the same solution as CVI$(F_{\hat{y}',\hat{z}}(x), K)$. 
\end{theorem}

{\bf Proof:} The proof is straightforward given the axioms of causal irrelevance \citep{pearl-book}. If $Y$ is causally irrelevant to $X$ given $Z$, then it follows that $P(x | \hat{y}, \hat{z}) = P(x | \hat{y}', \hat{z})$ for all $y, y', x, z$, namely if $\hat{z}$ is fixed, then changing the value of $y$ has no influence on the distribution of $x$. In this case, the mapping $F$ under the two intervention distributions remains identical. $\qed$

Now we examine the case when interventions do alter the solution to a causal VI, where our goal is to measure the change in solution in terms of properties of the ``untreated mapping $F_0$ and the ``treated" mapping $F_1$. 
\begin{theorem}
\label{sens-thm1}
Let the solution of the original ``untreated" CVI$(F_0,K)$be denoted by $x_0$, where $F_0$ is assumed to be strongly monotone, and (stochastically) Lipschitz, with $\mu$ being the coefficient in the strong monotonicity property. Given a causal intervention, the ``treated" CVI$(F_1, K)$ results in the modified solution vector $x_1$. Then it follows that
\begin{equation}
    \| x_1 - x_0 \| \leq \frac{1}{\mu} \| F_1(x_1) - F_0(x_1) \|
\end{equation}
\end{theorem}
{\bf Proof:} Since $x_0$ and $x_1$ solve the ``untreated" and "treated" causal VI's, respectively, it must follow that:
\begin{eqnarray*}
\langle F_0(x_0), y  - x_0 \rangle \geq 0, \ \ \forall y \in K \\
\langle F_1(x_1), y - x_1 \rangle \geq 0, \ \ \forall y \in K
\end{eqnarray*}
Substituting $y = x_1$ in the first equation above, and $y = x_0$ in the second equation, it follows that:
\[ \langle (F_1(x_1) - F_0(x_0)) , x_1 - x_0 \rangle \leq 0 \] 
Equivalently, we get
\[ \langle (F_1(x_1) - F_0(x_0) + F_0(x_1) - F_0(x_1)), x_1 - x_0 \rangle \leq 0 \]
Using the monotonicity property of $F_0$, we get:
\begin{eqnarray*}
\langle (F_1(x_1) - F_0(x_1)), x_0 - x_1 \rangle &\geq& \langle (F_0(x_0) - F_0(x_1)), x_0 - x_1 \rangle \\ &\geq& \mu \| x_0 - x_1 \|^2
\end{eqnarray*} 
from which the theorem follows immediately $\qed$

Interestingly, in the above analysis, we did not assume any property of the intervened causal VI $F_1$, other than it has a solution (meaning that $F_1$ should be continuous). The following corollaries follow directly from Theorem~\ref{sens-thm1}. 
\begin{theorem}
\label{sens-thm2}
Given the original ``untreated" causal VI CVI$(F_0,K)$, where $F_0$ is strictly monotone, and the intervened ``treated" causal VI CVI$(F_1, K)$, where the intervened mapping $F_1$ is continuous, but not necessarily monotone, if $x_0$ and $x_1$ denote the solutions to the original ``untreated" and causally intervened CVI, where $x_0 \neq x_1$, then it follows that:
\begin{eqnarray}
\langle (F_1(x_1) - F_0(x_1), x_1 - x_0 \rangle < 0 \\
\langle (F_1(x_1) - F_0(x_0), x_1 - x_0 \rangle \leq 0
\end{eqnarray}
\end{theorem}
 Here, we are bounding the causal intervention effect of $F_1 - F_0$ of the ``treated" vs. ``untreated" operator, whereas previously in the convergence analysis of Equation~\ref{2stepalg}, we were trying to bound the same operator's effect on two different parameter values. 
 The following theorem extends Theorem~\ref{sens-thm2} in showing that for partitionable CVI's, the effects induced by local causal interventions can be isolated.
\begin{theorem}
If a partitioned causal VI $CVI(F,K)$ is defined, where each component $F_i$ is a strongly monotone partitionable function,  $F^1_i$ denotes the causally intervened component function, and  $F^0_i$ is the ``untreated" function,  $x^0$ denotes the solution to the original ``untreated" CVI$(F_0,K)$ and $x^1$ denotes the solution to the ``treated" causally intervened CVI$(F_1,K)$ defined by the manipulated $F^1_i$ component functions, then 
\begin{equation}
\sum_{i=1}^m \langle (F^1_i(x^1_i) - F^0_i(x^1_i)), x^1_i - x^0_i \rangle  < 0
\end{equation}
\end{theorem}
{\bf Proof:} The proof follows readily from Theorem~\ref{sens-thm1}, Theorem~\ref{sens-thm2}, and Definition~\ref{partioned-cvi}. In particular, if $x^1$ is the solution of the intervened CVI$(F_1,K)$ and $x^0$ is the solution of the original CVI$(F_0,K)$, it follows that:
\[ \langle F_1(x^1) - F_0(x^1), x^1 - x^0 \rangle = \sum_{i=1}^m \langle F^1_i(x^1_i) - F^0_i(x^1_i), x^1_i - x^0_i \rangle \] 
Since the component functions $F_i$ are strongly monotone, the overall function $F$ is as well, and by applying Theorem~\ref{sens-thm1}, it follows that:
\[ \langle (F^1(x^1) - F^0(x^1)), x^1 - x^0 \rangle < 0 \] 
which immediately yields that
\[  \sum_{i=1}^m \langle (F^1_i(x^1_i) - F^0_i(x^1_i)), x^1_i - x^0_i \rangle < 0 \qed \]
If only a single component function $F_i$ is treated, then: 
\[  \langle (F^1_i(x^1_i) - F^0_i(x^1_i)), x^1_i - x^0_i \rangle < 0 \]
We can use these insights into designing an improved version of the two-step stochastic approximation algorithm given by Equation~\ref{2stepalg}. Instead of selecting random iterates to project on, we can instead prioritize those components $F^1_i$ that have been modified by the intervention.

\section{Summary and Future Work} 

In this paper we proposed the Universal Decision Model (UDM) framework, building on the core concept of information fields, suitably generalized to the formalism of category theory. We showed how information fields defined by decision objects in a UDM are associated with a finite topology, which can be exploited to facilitate hierarchical decomposition, as well as build homotopically invariant representations. We described a specific UDM category of causal variational inequalities, and showed how it can be used to solve causal inference problems in real-world complex network games.   We identified several universal properties, including information integration,  decision solvability,  and hierarchical abstraction.  Information integration is the process of consolidating data from heterogeneous sources, and its categorial foundation is built on forming products or limits in a category.  Abstraction enables simulating complex decision process by simpler processes through bisimulation morphisms, and its categorial foundation rests on forming quotients, co-products and co-limits. Finally, solvability refers to the requirement that a decision problem must have a unique solution defined by a fixed point equation, and it requires an order-preserving morphism across objects.  Much remains to be done in this research paradigm on universal decision making. We summarize a few topics for further research that extend the current paper. 

\subsection{Presheaf Representations}

We briefly described the Yoneda lemma, which specified how to construct universal representations for any object $c \in {\cal C}$ in a category based on covariant or contravariant functors. We can apply this approach to construct particular pre-sheaf ${\cal C}(-, c)$ representations of objects in the UDM category of MDPs, POMDPs, PSRS, and more generally intrinsic models. A detailed study of presheaf UDM representations is an important topic for future research. 

\subsection{Giri Monads}

In category theory, the usual way to model probability distributions is through {\em monads}, a topic we did not get into as it would take us far afield into category theory. Briefly, a Giri monad is the canonical monad structure for the category of all measurable spaces. A {\bf monoidal category} ${\cal C}$
 is a category that has a bifunctor $\otimes: {\cal C} \times {\cal C} \rightarrow {\cal C}$, along with an identity mapping, and several natural isomorphisms that define the associativity of the tensor product. 
 More formally, a monoid $(M, \mu, \eta)$ in a monoidal category ${\cal C}$ is an object $M$
 in ${\cal C}$ together with two morphisms (obeying the standard associativity and identity properties) that make use of the category’s monoidal structure: the associative binary operator $\mu: M \otimes M \rightarrow M$, and the identity $\eta: I \rightarrow M$. A {\bf monad} is often termed a ``monoid in the category of endofunctors", namely functors that map a category into itself. That is, consider the category of endofunctors whose objects are endofunctors and whose morphisms are natural transformations between them. This can be shown to define a monoidal category. 
 
 To link monads to probability distributions, recall that a measurable space $(X, {\cal F})$ is a set $X$ equipped with a $\sigma$-algebra ${\cal F}$. Recall also that a measure $\nu: X \rightarrow \mathbb{R}$
is a particular kind of set function from the $\sigma$-algebra to nonnegative real numbers. A measurable space completed with a measure $(X, {\cal F}, \nu)$ is called a measure space, and a measurable space completed with a probability measure is called a probability space. We have already previously defined measurable functions. We can now define the category ${\bf Meas}$ of measurable spaces, where the morphisms are simply the measurable mappings between them.  For any specific measurable space $M$, we can define 
the space of all possible probability measures that could be placed on it as $\Xi(M)$. Note that $\Xi(M)$
 is itself a space of measures - that is, a space in which the points themselves are probability measures. As a probability measure, any element of $\Xi(M)$ is a function from measurable subsets of $M$ to the interval $[0,1]$ in $\mathbb{R}$.  A key area for future work is to study UDMs defined over Giri monads.

\subsection{Kan Extensions of Intrinsic Models} 

It is well known in category theory that ultimately every concept, from products and co-products, limits and co-limits, and ultimately even the Yoneda embeddings, can be derived as special cases of the Kan extension \citep{riehl}. Another topic for future work is to apply this powerful technique in the analysis of intrinsic models. We briefly define the concept of Kan extensions below. Kan extensions intuitively are a way to approximate a functor ${\cal F}$ so that its domain can be extended from ${\cal C}$ to ${\cal D}$. In other words, Kan extensions are a way of taking two functors and constructing a third functor to make a diagram commute. Because it may be impossible to make commutativity work in general, Kan extensions rely on natural transformations to make the extension be the best possible approximation to ${\cal F}$ along ${\cal K}$. 

\begin{definition}
A {\bf left Kan extension} of a functor $H: {\cal C} \rightarrow {\cal E}$ along $F$, another functor $F: {\cal C} \rightarrow {\cal D}$, is a functor $\mbox{Lan}_F H: {\cal D} \rightarrow {\cal E}$ with a natural transformation $\eta: H \rightarrow \mbox{Lan}_F H\circ F$ such that for any other such pair $(G: {\cal D} \rightarrow {\cal E}, \gamma: H \rightarrow G K)$, $\gamma$ factors uniquely through $\eta$. In other words, there is a unique natural transformation $\alpha: \mbox{Lan}_F \implies G$. \\
%
\begin{center}
\begin{tikzcd}[row sep=2cm, column sep=2cm]
\mathsf{C}  \ar[dr, "K"', ""{name=K}]
            \ar[rr, "F", ""{name=F, below, near start, bend right}]&&
\mathsf{E}\\
& \mathsf{D}    \ar[ur, bend left, "\text{Lan}_KF", ""{name=Lan, below}]
                \ar[ur, bend right, "G"', ""{name=G}]
%
\arrow[Rightarrow, "\exists!", from=Lan, to=G]
\arrow[Rightarrow, from=F, to=K, "\eta"]
\end{tikzcd}
\begin{tikzcd}[row sep=huge, column sep=huge] 
 \mathcal{C} \arrow[dr, "\mathcal{F}"'{name=F}] 
 \arrow[rr, "\mathcal{H}", ""{name=H, below}] && \mathcal{E} \\ 
 & |[alias=D]| \mathcal{D} \arrow[ur, swap, dashed,
 "\operatorname{Lan}_{\mathcal{F}}\mathcal{H}"] 
 \arrow[Rightarrow, from=H, to=D, "\eta",shorten >=1em,shorten <=1em] 
\end{tikzcd} 
\end{center}
\end{definition}

A key challenge for future work is to explore approximation of functors from the UDM category to other categories using Kan extensions. 

\subsection{Applications} 

 We introduced the Universal Decision Model (UDM), a broad overarching framework for decision making that integrates a number of well-studied modalities, including causal inference, decentralized stochastic control and reinforcement learning, and multiplayer games in network economics. The UDM model uses category theory to contruct a universe of decision making objects, which are related by bisimulation morphisms. The information field representation defines the knowledge available to each decision maker, and induces a finite topology on the space of agents. The topology of subsystems allows hierarchical decomposition of complex networks of agents, and we showed how homotopically equivalent systems can be formulated using algebraic topology.  The next step is to articulate how specific applications can be solved in this paradigm, and design effective algorithms for this purpose. Applications naturally would require making concrete choices on the particular types of problem classes involved (e.g., team, classical, sequential etc.), and particular types of temporal ordering (e.g., linear, partial), and  subsystem design using specific information fields. 







\end{document}